\documentclass[11pt]{article}

\usepackage[final]{acl}

\usepackage{times}
\usepackage{latexsym}

\usepackage[T1]{fontenc}
\usepackage[utf8]{inputenc}

\usepackage{microtype}

\usepackage{inconsolata}

\usepackage{graphicx}
\title{Debiasing Large Language Models via Adaptive Causal Prompting with Sketch-of-Thought}

\author{
  \textbf{Bowen Li\textsuperscript{1}},
  \textbf{Ziqi Xu\textsuperscript{1}},
  \textbf{Jing Ren\textsuperscript{1}},
  \textbf{Renqiang Luo\textsuperscript{2}},
\\
  \textbf{Xikun Zhang\textsuperscript{1}},
  \textbf{Xiuzhen Zhang\textsuperscript{1}},
  \textbf{Yongli Ren\textsuperscript{1}},
  \textbf{Feng Xia\textsuperscript{1}}
\\
\\
  \textsuperscript{1}RMIT University, Australia,
  \textsuperscript{2}Jilin University, China
\\
 \small{
   \textbf{Correspondence:} \href{mailto:ziqi.xu@rmit.edu.au}{ziqi.xu@rmit.edu.au}
 }
}

\usepackage{graphicx}
\usepackage{multirow}
\usepackage{subcaption}
\usepackage{booktabs}
\usepackage{array}
\usepackage{enumitem}
\usepackage{siunitx}
\usepackage{makecell}
\usepackage{amsmath}
\usepackage{amssymb} % also loads amsfonts
\usepackage{pifont}
\usepackage{algpseudocode}
\newtheorem{definition}{Definition}
\newtheorem{theorem}{Theorem}
\usepackage[most]{tcolorbox}
\usepackage{algorithm}
\usepackage{algcompatible}
\usepackage{xcolor, soul} % (if not already loaded)
\sethlcolor{yellow}
\soulregister\cite7
\soulregister\ref7

\newcommand{\indep}{\mathrel{\perp\!\!\!\perp}}

\newcommand{\makepromptbox}[2]{
\begin{tcolorbox}[colback=brown!5!white,
                  colframe=brown!50!black,
                  colbacktitle=brown!75!black,
                  title=#1,
                  width=\columnwidth,
                  boxrule=0.8mm,
                  breakable]
\small #2
\end{tcolorbox}
}

\begin{document}
\maketitle
\begin{abstract}
Despite notable advancements in prompting methods for Large Language Models (LLMs), such as Chain-of-Thought (CoT), existing strategies still suffer from excessive token usage and limited generalisability across diverse reasoning tasks. To address these limitations, we propose an \textbf{A}daptive \textbf{C}ausal \textbf{P}rompting with \textbf{S}ketch-of-Thought (\textbf{ACPS}) framework, which leverages structural causal models to infer the causal effect of a query on its answer and adaptively select an appropriate intervention (i.e., standard front-door and conditional front-door adjustments). This design enables generalisable causal reasoning across heterogeneous tasks without task-specific retraining. By replacing verbose CoT with concise Sketch-of-Thought, ACPS enables efficient reasoning that significantly reduces token usage and inference cost. Extensive experiments on multiple reasoning benchmarks and LLMs demonstrate that ACPS consistently outperforms existing prompting baselines in terms of accuracy, robustness, and computational efficiency. The source code can be found at~\url{https://aisuko.github.io/acps/}.
\end{abstract}

\section{Introduction}
Large Language Models (LLMs) play a central role in Natural Language Processing (NLP), achieving state-of-the-art results across a wide range of tasks, from open-domain question answering to multi-step logical reasoning~\cite{Brown2020GPT3}. Building on this success, prompt-based methods extend LLM capabilities without requiring full model retraining. For example, In-Context Learning (ICL)~\cite{Xie2022BayesICLR} introduces example demonstrations directly into the prompt, enabling LLMs to generalise from just a few instances. To support more complex reasoning, Chain-of-Thought (CoT) prompting elicits step-by-step inference, substantially improving performance on multi-hop and logical tasks~\cite{Wei2022CoT}.

Despite recent advances, current prompting strategies exhibit two critical shortcomings. First, excessive token generation remains a major issue: CoT prompts often produce unnecessarily lengthy or redundant reasoning chains, with hundreds of tokens per query, which reduces efficiency and increases inference cost~\cite{Xu2025ChainOfDraft}. While methods such as Chain-of-Draft (CoD)~\cite{Xu2025ChainOfDraft} and Sketch-of-Thought (SoT)~\cite{Aytes2025SketchOfThought} attempt to alleviate this, they typically lag behind CoT in accuracy. Second, unfaithful reasoning emerges from internal model biases, where LLMs may rationalise a predisposed answer instead of performing genuine inference. Subtle bias cues in prompts can lead to plausible yet factually incorrect CoTs, resulting in accuracy drops of up to 36\% on complex reasoning tasks~\cite{Turpin2023Unfaithful}. Such reasoning failures undermine trust in applications that require faithful and verifiable explanations, such as scientific question answering and commonsense inference.

Recent research has explored the integration of causal inference with LLMs as a promising direction for addressing the aforementioned challenges. By incorporating causal principles such as standard front-door adjustment and instrumental variable, it becomes possible to mitigate bias caused by unobserved confounders, which is often interpreted as internal bias in LLMs. These unobserved confounders can introduce spurious correlations between the query and the answer, leading to unfaithful or misleading outputs. For example, Causal Prompting~\cite{zhang2024causal} estimates the causal effect of a query its answer by controlling for intermediate reasoning steps such as CoT. Instead of relying on majority voting across multiple reasoning paths, this method selects the answer with the highest estimated causal effect, thereby improving both accuracy and interpretability.

However, despite its potential, existing causality-based prompting methods rely on strong assumptions, including the identifiability conditions required by front-door adjustment and the validity of instrumental variables. These assumptions may not hold consistently across different NLP tasks, limiting the generalisability of such methods in practice. Furthermore, the reliance on CoT often leads to verbose outputs, which increase both token usage and inference cost. Thus, there is a pressing need for a causality-based prompting framework that mitigates internal biases in LLM reasoning by adaptively selecting an appropriate intervention for different NLP tasks, while efficiently reducing token usage without compromising performance.

\begin{figure}[t]
    \centering
    \includegraphics[width=0.48\textwidth]{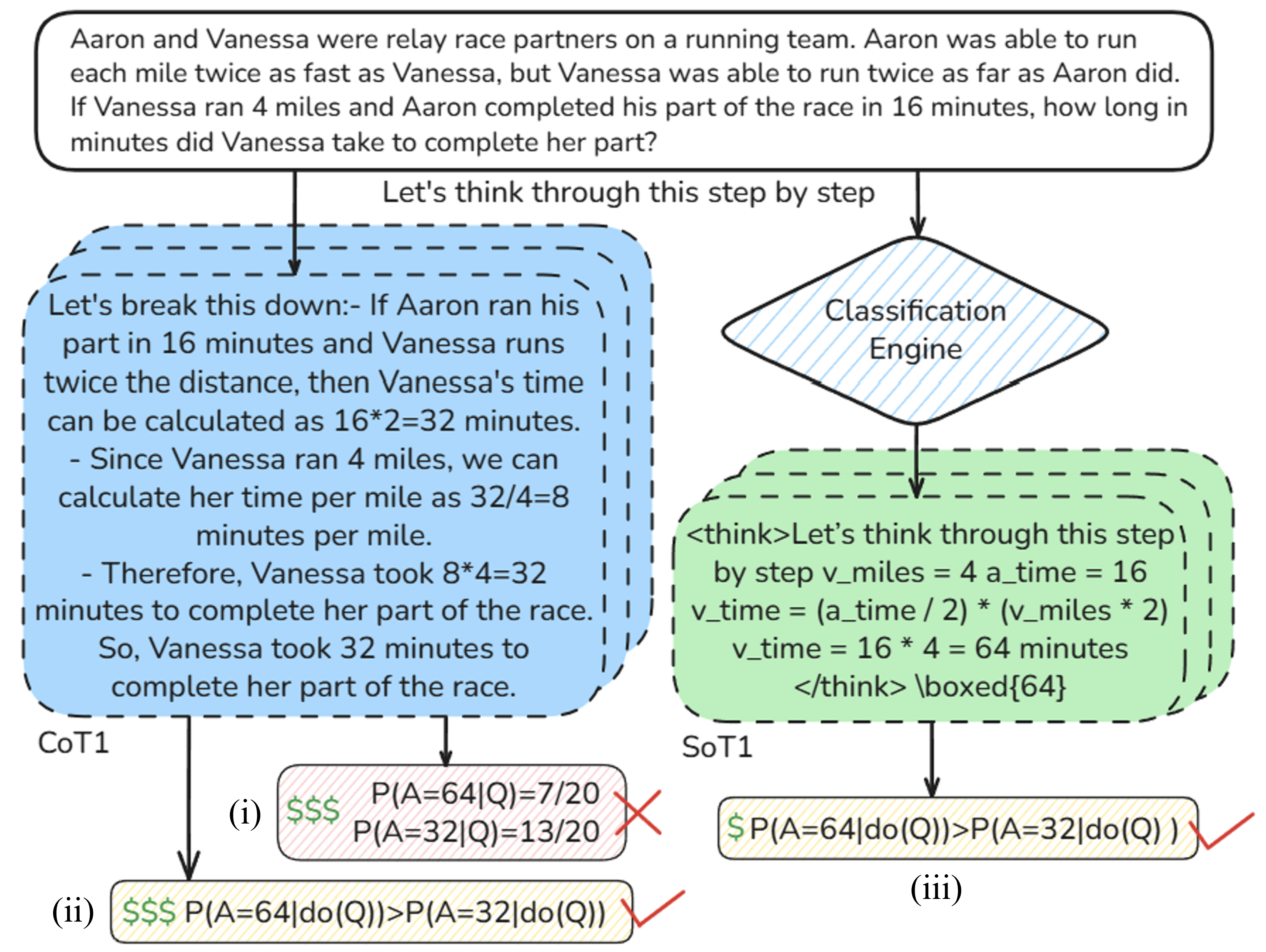}
    \caption{An example from GPT-3.5-turbo on the GSM8K dataset. Left: (i) Recent non-causal prompting methods often amplify internal bias through majority voting. (ii) Some causality-based prompting methods mitigate this bias but rely on verbose CoT, leading to high token usage and inference cost. Right: (iii) The proposed framework uses SoT instead of CoT and selects the answer based on the highest estimated causal effect, yielding the correct result.}
    \label{fig:figure1}
\end{figure}

In this work, we propose an \textbf{A}daptive \textbf{C}ausal \textbf{P}rompting with \textbf{S}ketch-of-Thought (\textbf{ACPS}) framework to enhance both the generalisability and efficiency of debiasing LLMs. ACPS adaptively applies standard or conditional front-door adjustment depending on the characteristics of each NLP task, supported by a classification engine. To improve inference efficiency, it replaces verbose CoT with concise SoT, significantly reducing token usage and computational cost. 
%As shown in Figure~\ref{fig:figure1}, ACPS produces the correct answer on a representative GSM8K example, whereas non-causal prompting methods suffer from internal bias and causality-based prompting methods relying on verbose CoT are limited by inefficiency. 
As shown in Figure~\ref{fig:figure1}, ACPS produces the correct answer on a representative GSM8K example. In contrast, non-causal prompting methods suffer from internal bias, while causality-based prompting methods that rely on verbose CoT are constrained by inefficiency.
This highlights how our adaptive formulation and concise reasoning traces contribute to improved accuracy and efficiency in inference. The main contributions of this paper are as follows:
\begin{itemize}
    \item We propose a novel framework, ACPS, for debiasing LLMs via adaptive causal prompting. This model-agnostic mechanism selects an appropriate intervention based on task characteristics, thereby overcoming the limitations of fixed prompting and improving generalisability across diverse reasoning tasks.

    \item To the best of our knowledge, ACPS is the first framework that integrates both standard and conditional front-door adjustments with SoT. This integration significantly reduces token usage and inference cost, while preserving high reasoning accuracy.

    \item We validate our framework through extensive experiments across multiple LLMs and reasoning benchmarks, demonstrating consistent improvements in accuracy, efficiency, and robustness over existing prompting baselines.
\end{itemize}

\section{Preliminaries}
In this section, we review the fundamental concepts of causality and sketch-of-thought.

\subsection{Structural Causal Model}
We adopt the structural causal model (SCM)~\cite{pearl2016causal}, in which each endogenous variable is determined by a deterministic function of its parent variables and an independent exogenous noise term. The causal structure is represented by a directed acyclic graph (DAG), where nodes correspond to random variables and directed edges denote direct causal relationships. Exogenous variables are assumed to be mutually independent, allowing the joint distribution to factorise according to the graph structure. We assume the Markov condition~\citep{Pearl2009Causality}, which states that each variable is conditionally independent of its non-effects given its direct causes, and the faithfulness assumption~\citep{spirtes2000causation}, which asserts that all and only those conditional independencies implied by the DAG are present in the observed distribution. Under these assumptions, d-separation~\citep{Pearl2009Causality} can be used as a criterion to determine conditional independence relationships. Interventions, denoted by \( do(X = x) \), modify the structural equation for \( X \), producing a new distribution that reflects the causal effect of the intervention. For formal definition and further details, see~\citep{Pearl2009Causality}. 

\begin{figure}
    \centering
    \begin{subfigure}[t]{0.12\textwidth}
        \includegraphics[width=\textwidth]{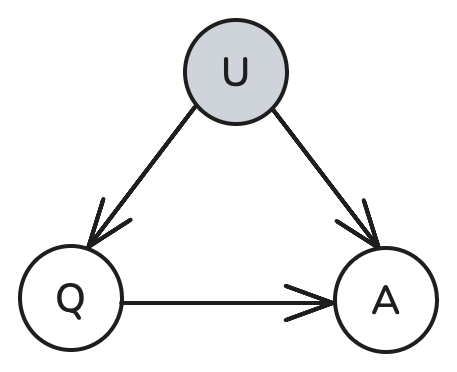}
        \caption{}
        \label{fig2:a}
    \end{subfigure}
    \hfill
    \begin{subfigure}[t]{0.17\textwidth}
        \includegraphics[width=\textwidth]{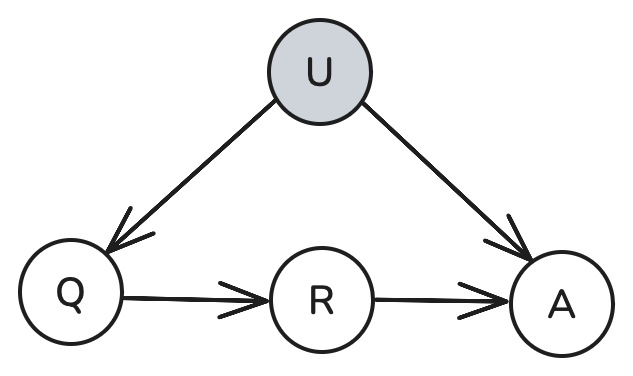}
        \caption{}
        \label{fig2:b}
    \end{subfigure}
    \hfill
    \begin{subfigure}[t]{0.17\textwidth}
        \includegraphics[width=\textwidth]{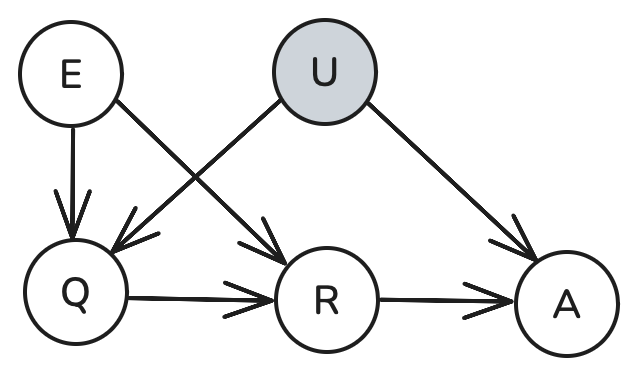}
        \caption{}
        \label{fig2:c}
    \end{subfigure}
    \caption{Three SCMs illustrate different modes of reasoning in LLMs: (a) direct prompt-to-answer reasoning; (b) causality-based prompting for tasks without external knowledge, such as Causal Prompting~\cite{zhang2024causal}; and (c) causality-based prompting for tasks with external knowledge. Both (b) and (c) are integrated into ACPS. In all SCMs, \( Q \) denotes the query, \( R \) denotes the reasoning process (SoT or CoT), \( A \) denotes the answer, \( U \) denotes the unobserved confounder, and \( E \) denotes the external knowledge.}
    \label{fig:total}
\end{figure}

For direct prompt-to-answer reasoning, the conceptual-level SCM is illustrated in Figure~\ref{fig2:a}. The query \( Q \), which includes both demonstrations and test examples provided to the LLM, leads to the answer \( A \). Although the direct causal effect from \( Q \) to \( A \) is represented as \( Q \to A \), LLMs often internalise biases from large-scale pre-training corpora~\cite{Ding2023PEFT, Aghajanyan2021ID,ren2025causal}. To account for these biases, we introduce an unobserved variable \( U \) that influences both \( Q \) and \( A \). The presence of \( U \) creates a spurious association between \( Q \) and \( A \), which can result in biased or incorrect answers during inference.

\subsection{Sketch-of-Thought}
SoT is a prompting framework that rethinks how LLMs express reasoning, addressing the verbosity of CoT~\cite{Aytes2025SketchOfThought}. Rather than full-sentence explanations, SoT elicits concise sketches, which are abridged representations that capture essential logical structure while omitting details and thereby reducing token usage. The notion of a sketch, drawn from cognitive science~\cite{Goel1995}, denotes a symbolic intermediate form that preserves core reasoning while abstracting away irrelevance. By combining cognitive inspiration with linguistic conciseness, SoT produces compact and interpretable reasoning traces. The SoT template is provided in Appendix~\ref{appendix:promtp_templates}.

\section{Methodology}
\begin{figure*}[t]
    \centering
    \includegraphics[width=0.99\textwidth]{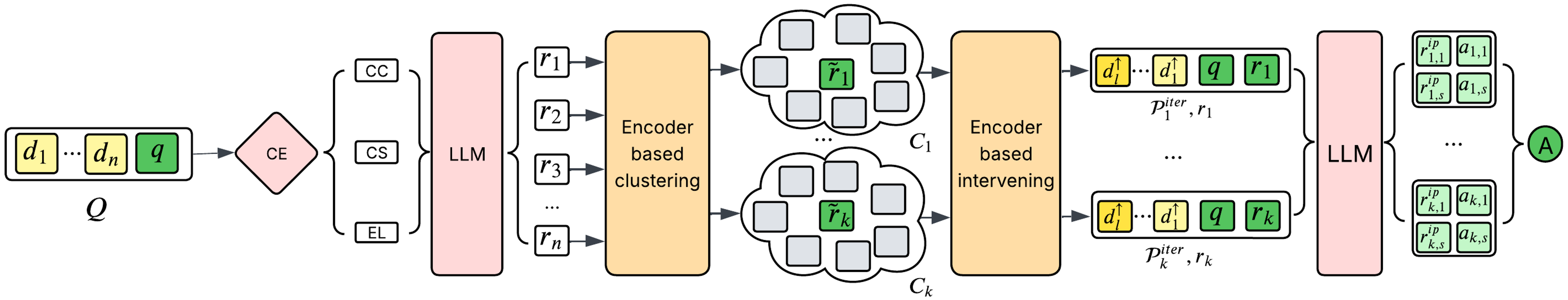}
    \caption{Overall architecture of ACPS. Given an input $Q$ comprising the demonstration examples $[d_1, \ldots, d_n]$ and the test query $q$, a classification engine (CE) determines an appropriate intervention. The LLM generates $M$ diverse SoTs, which are embedded and clustered into $K$ groups. For each cluster representative, optimal demonstrations are selected via an encoder-based intervention algorithm to form updated prompts $\mathcal{P}^{\text{iter}}_{k}$. The LLM is then queried $S$ times per prompt, and the final answer is selected as the one associated with the highest estimated causal effect.}
    \label{fig:architecture}
\end{figure*}

In this section, we first introduce the causal principles, including the standard and conditional front-door criteria. We then describe the mechanism for adaptively selecting an appropriate intervention, followed by the estimation process for each component. Finally, we derive the overall objective function that quantifies the causal effect of the query on its answer. The overall architecture of ACPS is shown in Figure~\ref{fig:architecture}.

\subsection{Causal Principles}
A key approach to handling unobserved confounders is the front-door adjustment~\cite{Pearl2009Causality}. Unlike the back-door criterion, the front-door criterion can isolate causal pathways even when confounders are unobserved. This property makes it particularly suitable for reducing internal biases in LLMs, where unobserved confounders may exist but intermediate variables (i.e., SoT) are observable and controllable. Below, we present the standard front-door criterion and describe its adaptation for prompt-based debiasing.

\begin{definition}[Standard Front-Door Criterion]
    \label{def:FD}
    A set of variables \( Z_{\text{SFD}} \) is said to satisfy the standard front-door criterion with respect to an ordered variable pair \( (Q, A) \) in a DAG \( \mathcal{G} \) if the following conditions are met: (1) \( Z_{\text{SFD}} \) intercepts every directed path from \( Q \) to \( A \); (2) there is no unblocked back-door path from \( Q \) to \( Z_{\text{SFD}} \); (3) all back-door paths from \( Z_{\text{SFD}} \) to \( A \) are blocked by \( Q \).
\end{definition}

The standard front-door criterion offers a theoretical basis for identifying causal effects despite unobserved confounders. In LLMs, this supports using SoT reasoning as a valid front-door variable to estimate the causal effect of prompts on final answers. As shown in Figure~\ref{fig2:b}, \(R\) meets the standard front-door conditions for the causal effect of \(Q\) on \(A\). This allows the causal effect \( P(A \mid do(Q)) \) to be decomposed into two parts: the effect of \(Q\) on \(R\), and the effects of \(R\) on \(A\) conditioned on \(Q\). Formally, the front-door adjustment formula is:
\begin{equation}
\label{final22}
    \centering
    P(A \mid do(Q)) = \sum_{r,~q} \underbrace{P(r \mid Q)}_{\text{\ding{172}}} \underbrace{P(A \mid r, q)}_{\text{\ding{173}}}.
\end{equation}

However, Figure~\ref{fig2:c} presents a distinct class of reasoning tasks in which an LLM receives a query \(Q\), generates a SoT \(R\), and produces an answer \(A\). In this setting, external knowledge \(E\) influences both \(Q\) and \(R\), while an unobserved confounder \(U\) introduces spurious correlations that bias causal effect estimation. As this scenario falls outside the scope of the standard front-door criterion, we adopt a conditional front-door adjustment~\cite{Xu2024CFDICLR} to accurately estimate \( P(A \mid do(Q)) \) and identify the most reliable answer. The formal criterion is defined as follows:

\begin{definition}[Conditional Front-Door Criterion]
    \label{def:FD1}
    A set of variables \( Z_{\text{\rm CFD}} \) satisfies the \textit{conditional front-door criterion} relative to an ordered pair \( (Q, A) \) in a DAG \( \mathcal{G} \) if the following conditions hold:  
    (1) \( Z_{\text{\rm CFD}} \) intercepts all directed paths from \( Q \) to \( A \);  
    (2) there exists a set of variables \( W \) such that all back-door paths from \( Q \) to \( Z_{\text{\rm CFD}} \) are blocked by \( W \);  
    (3) all back-door paths from \( Z_{\text{\rm CFD}} \) to \( A \) are blocked by \( Q \cup W \).
\end{definition}

As shown in Figure~\ref{fig2:c}, \(R\) meets all the requirements of the conditional front-door criterion for the pair \((Q,A)\), with the external knowledge variable \(E\) acting as the conditioning set \(W\). Thus, \(R\) serves as a valid conditional front-door adjustment variable to identify the causal effect of \(Q\) and \(A\). 

In our framework, \(Q\) represents the fixed query during reasoning. Since no intervention is applied to \(Q\), it is considered a constant instead of a random variable. Therefore, the term \( P(q \mid e) \) and the summation over \(q\) can be removed, simplifying the causal effect expression as follows:

{\small
\begin{equation}
    \label{final33}
    \centering
    P(A \mid do(Q)) = \sum_{r,~q,~e}  \underbrace{P(r \mid Q, e)}_{\text{\ding{172}}} \underbrace{P(A \mid r,q,e)}_{\text{\ding{173}}}\underbrace{P(e)}_{\text{\ding{174}}}
\end{equation}}

We decompose the causal effect of the query \( Q \) on its answer \( A \) using two formulations based on the nature of the task. For tasks without external knowledge \( E \), we adopt the standard front-door formulation as shown in Eq.~\ref{final22}, which contains two components that can be independently estimated. For tasks involving external knowledge, we apply the conditional front-door formulation shown in Eq.~\ref{final33}, which extends the decomposition to three components by incorporating conditioning on \( E \). The derivation is provided in Appendix~\ref{appendix:derivation}.

In the following Sections~\ref{2}, \ref{3}, and \ref{1}, we detail how each component is estimated in practice. We focus primarily on the conditional front-door setting, as its estimation procedure generalises naturally to the standard front-door case by simply omitting the conditioning on \( E \).

\subsection{Classification Engine}
To adaptively determine an appropriate intervention for each query, we directly adopt a fine-tuned DistilBERT classification engine~\cite{Sanh2019DistilBERT} from previous work~\cite{Aytes2025SketchOfThought} without further training. The engine assigns each query to one of three reasoning paradigms: Conceptual Chaining (CC), Chunked Symbolism (CS), or Expert Lexicons (EL). Given a question \(x\), the classifier outputs logits \(z_c\) for each class \(c \in \{\text{CC}, \text{CS}, \text{EL}\}\), which are transformed into a probability distribution over the classes using the softmax function:

{\small\begin{equation}
    P(c \mid x) = \frac{\exp(z_c)}{\sum_{c' \in C} \exp(z_{c'})}, \quad C = \{\text{CC}, \text{CS}, \text{EL}\},
\end{equation}}where \( P(c \mid x) \) denotes the probability that the input question \( x \) belongs to class \( c \), \( z_c \) is the logit associated with class \( c \), and \( C \) is the set of candidate reasoning paradigms. The variable \( c' \) in the denominator is a dummy index that iterates over all classes in \( C \).

Then, the argmax operation selects the reasoning paradigm with the highest probability:
\begin{equation}
c^* =  \arg\max_{c\in \{\text{CC}, \text{CS}, \text{EL}\}} P(c|x),
\end{equation}
where \(c^*\) denotes the predicted reasoning category with the maximum likelihood.

Each paradigm corresponds to a distinct type of reasoning task. CC includes tasks that require synthesising multiple pieces of contextual or external information. These tasks are handled using Eq.~\ref{final33}, which applies the conditional front-door adjustment. CS consists of tasks such as arithmetic or algebraic problem solving, where the reasoning involves symbolic steps without reliance on external knowledge. These tasks are addressed using Eq.~\ref{final22}, corresponding to the standard front-door adjustment. EL covers tasks involving commonsense inference and factual verification. For EL, the selection between Eq.~\ref{final22} and Eq.~\ref{final33} depends on the availability of external knowledge: if external information is present, the conditional version is used; otherwise, the standard version applies.

\subsection{Estimating Reasoning Trace Distribution}
\label{2}
We estimate the causal effect between the query \(Q\), the external knowledge \(E\), and the SoT \(R\). To address challenges such as inaccessible LLM output probabilities and limited diversity in SoTs, we generate SoTs by varying the temperature parameter from \(0.0\) to \(2.0\) in increments of \(0.25\), resulting in a diverse set \(R = [r_{1}, r_{2}, \dots, r_{m}]\). To maximise the diversity of the generated SoTs, we compute their embeddings using a pre-trained LLM, Sentence-BERT~\cite{reimers-2019-sentence-bert}, as the encoder, producing embeddings \(\tilde{R} = [\tilde{r}_{1}, \tilde{r}_{2}, \dots, \tilde{r}_{m}]\). These embeddings are then clustered using the K-Means algorithm~\cite{Ikotun2023KMeans} to partition them into \(K\) clusters:
\begin{equation}
    \{C_{1}, \dots, C_{k}\} = \mathrm{K\text{-}means}(\tilde{r}_{1}, \dots, \tilde{r}_{m}),
\end{equation}
where \(C_{k}\) denotes the \(k\)-th cluster in the result, and \(K\) is the total number of clusters.

For each cluster, we select the SoT that is closest to the cluster centroid, resulting in a set of \(K\) representative SoTs to be used in subsequent causal analysis:
\begin{equation}
\label{equation_r_k}
    r_{k} = \mathrm{Center}(C_{k}).
\end{equation}

We estimate the conditional probability \( P(r \mid Q, e) \) based on the size of each cluster as follows:
\begin{equation}
\label{222}
    P(r \mid Q, e) \approx \frac{|C_{k}|}{M},
\end{equation}
where \(|C_k|\) denotes the number of SoTs in the \(k\)-th cluster and \(M\) is the total number of generated SoTs.

\subsection{Estimating Final Answer Probability}
\label{3}
In this section, we estimate the causal effect of the answer \(A\) produced by the LLM, conditioned on the query \(Q\), the external knowledge \(E\), and the SoT \(R\). The main challenge arises from the virtually unlimited value space of both \(Q\) and \(E\). To address this, we apply the encoder-based Normalised Weighted Geometric Mean (NWGM) approximation~\cite{xu2015show_attend_tell} in ICL prompting interventions. We construct a fixed-size demonstration set \(D = \{ d_n = (q_n, e_n, r^{\text{wrong}}_n, r^{\text{correct}}_n) \}_{n=1}^{N}\), where \(q_n\) and \(e_n\) represent the question and context of the \(n\)-th sample, respectively, and \(r^{\text{wrong}}_n\) and \(r^{\text{correct}}_n\) correspond to incorrect and correct SoTs for that sample.

We use the encoder to compute the embedding \(\tilde{r}_{k}\) of the \(k\)-th SoT \(r_{k}\), as selected in Eq.~\ref{equation_r_k} from Section~\ref{2}. Next, we calculate the similarity between \(\tilde{r}_{k}\) and each sample in the ICL demonstration set to improve performance in in-context learning~\cite{margatina2023active}. Finally, we rank the ICL demonstrations based on these similarity scores to determine the relevance of each sample, as follows:
\begin{equation}
\label{re_rank}
    \{d^{\uparrow}_{n}\}^{N}_{n=1} = \mathrm{Sort}(D, \tilde{r}_{k}, \{\tilde{d}_{n}\}^{N}_{n=1}),
\end{equation} where \(d^{\uparrow}_{n}\) denotes the sorted demonstration examples, and \(\mathrm{Sort}(\cdot)\) refers to the process of ranking samples using a cosine similarity function \(\cos(\cdot,\cdot)\). The demonstrations are ordered such that \(\cos(\tilde{r}_{k}, \tilde{d}_{i}) \geq \cos(\tilde{r}_{k}, \tilde{d}_{j})\) for all \(i < j\).

Then the \(L\) most similar samples from the ICL demonstration set are selected to form the prompt, where \(L \ll N\). The most similar samples are placed closest to the test query, as this ordering has been shown to better support the encoder-based NWGM algorithm in improving SoT quality through practical experiments. For each SoT \(r_k\) of a test sample, the final prompt after intervention is constructed as:
\begin{equation}
\label{intervention_r_k}
    \mathcal{P}^{\text{iter}}_{k} = [d^{\uparrow}_{l}, \dots, d^{\uparrow}_{1}, q^{\text{test}}].
\end{equation}

Subsequently, we query the LLM \(S\) times using the prompt \(\mathcal{P}^{\text{iter}}_{k}\) and SoT \(r_k\), generating \(S\) answers and corresponding improved SoTs. The probability \(P(A \mid r, q, e)\) is then estimated as follows:
\begin{equation}
\label{333}
    P(A \mid r, q, e) \approx \frac{1}{S} \sum_{s=1}^S \mathbb{I}(A = a_{k,s}),
\end{equation} where \(\mathbb{I}(\cdot)\) is the indicator function that returns 1 if the generated answer \(a_{k,t}\) matches the expected \(A\), and 0 otherwise.

\subsection{Estimating External Knowledge Distribution}
\label{1}
We maintain the external knowledge \(E\) fixed and integrate it directly with the query \(Q\). This integration provides the necessary context for reasoning and allows the conditional front-door framework to estimate causal effects without explicitly generating or manipulating additional knowledge.

Specifically, each \(Q\) is combined with its corresponding \(E\) to form a unified input that guides the reasoning process. This joint representation enables the conditional front-door adjustment to capture the underlying causal relationships within multi-hop language processing tasks. By leveraging contextual information directly, the proposed method simplifies the reasoning pipeline while preserving essential dependencies. The distribution of \(E\) is assumed to factorise as:
\begin{equation}
    P(E) = \prod_{i} P(e_i),
\end{equation} where \(e_i\) denotes an individual element within \(E\).

\subsection{Objective Function for ACPS}
\label{4}
Building on the results from Sections~\ref{2}, \ref{3}, and~\ref{1}, the final objective can be estimated as follows:

\vspace{-0.5cm}
{\small\begin{align}
\label{666}
    &P(A \mid do(Q)) 
    = \sum_{r,~e,~q} P(r \mid Q, e) \cdot P(A \mid r, q, e) \cdot P(e) \notag \\ 
    &~\quad\approx \sum_{k=1}^{K} \left[ \prod_{i=1} P(e_i) \right] \cdot \frac{|C_k|}{M} \cdot \frac{1}{S} \sum_{s=1}^{S} \mathbb{I}(A = a_{k,s})
\end{align}}

This equation estimates the causal effect between the query \( Q \) and the answer \( A \), enabling the selection of the answer with the highest estimated causal effect as the final unbiased output. The complete learning procedure is detailed in Appendix~\ref{appendix:Algorithm}.

\begin{table*}[t]
    \small
    \centering
    \setlength{\tabcolsep}{8pt}
    \renewcommand{\arraystretch}{0.8}
    \begin{tabular}{l|l
                |cc   % GSM8K & MATH Acc
                |cc   % CommonsenseQA & StrategyQA Acc
                |cc   % HotpotQA EM/F1
                |cc   % MuSiQue EM/F1
                |c}   % FEVER Acc
    \toprule
    
    {} & {Method} & {GSM8K} & {MATH}  & {ComQA} & {StrQA} 
      & \multicolumn{2}{c}{{HotpotQA}}
      & \multicolumn{2}{c|}{{MuSiQue}}
      & {FEVER} \\
    \cmidrule(lr){3-3}
    \cmidrule(lr){4-4}
    \cmidrule(lr){5-5}
    \cmidrule(lr){6-6}
    \cmidrule(lr){7-8}
    \cmidrule(lr){9-10}
    \cmidrule(lr){11-11}
    & & {Acc}~$\uparrow$ & {Acc}~$\uparrow$ & {Acc}~$\uparrow$ & {Acc}~$\uparrow$ 
             & {EM}~$\uparrow$ & {F1}~$\uparrow$ 
             & {EM}~$\uparrow$ & {F1}~$\uparrow$ & {Acc}~$\uparrow$ \\
    \midrule
    
    %===== Ministral-3B =====%
    \multirow{8}{*}{\rotatebox[origin=c]{90}{Ministral-3B}}
      & ICL     & 14.00 & 4.41 & 20.00 & 45.34 & 26.92 & 42.06 & 16.22 & 31.81 & 29.41 \\
      & CoT     & 36.09 & 36.29 & 44.44 & 51.72 & 31.58 & 47.20 & 31.05 & 41.30 & 31.58 \\
      & CoT-SC  & 42.86 & 40.12 & 38.73 & 55.10 & 33.33 & 49.16 & 40.00 & 50.02 & 41.67 \\
      & SoT     & 35.36 & 36.75 & 43.45 & 52.63 & 31.19 & 46.37 & 30.47 & 40.12 & 31.15 \\
      & CAD     & — & — & — & 54.24 & 35.23 & 49.01 & 41.25 & 52.65 & 39.65 \\
      & DeCoT   & — & — & — & 53.15 & 35.65 & 50.21 & 40.75 & 53.66 & 41.58 \\
      & CP      & \textbf{63.16} & 33.73 & 50.00 & 55.56 & 38.03 & 48.98 & 43.48 & 51.31 & 54.55 \\
      & ACPS    & 61.90 & \textbf{37.93} & \textbf{53.67} & \textbf{67.80} & \textbf{50.00} & \textbf{66.67} & \textbf{51.72} & \textbf{60.65} & \textbf{60.00} \\
    \midrule
    
    %===== LLaMA-3 =====%
    \multirow{8}{*}{\rotatebox[origin=c]{90}{LLaMA-3}}
      & ICL     & 18.76 & 18.75 & 28.45 & 56.91 & 32.55 & 43.18 & 32.55 & 43.18 & 45.82 \\
      & CoT     & 38.10 & 40.35 & 61.73 & 52.80 & 39.16 & 50.20 & 38.64 & 48.23 & 52.45 \\
      & CoT-SC  & 30.77 & 42.08 & 57.14 & 64.29 & 40.88 & 55.94 & 38.46 & 51.34 & 53.02 \\
      & SoT     & 37.78 & 40.83 & 60.76 & 52.04 & 39.17 & 54.44 & 42.73 & 47.24 & 52.20 \\
      & CAD     & — & — & — & 69.23 & 53.08 & 63.95 & 40.10 & 53.33 & 53.55 \\
      & DeCoT   & — & — & — & 70.95 & 54.63 & 64.52 & 41.23 & 54.56 & 57.52 \\
      & CP      & 69.67 & 46.67 & 60.10 & 72.10 & 55.00 & 69.98 & 44.94 & 53.15 & 63.64 \\
      & ACPS    & \textbf{79.71} & \textbf{46.80} & \textbf{63.64} & \textbf{73.03} & \textbf{56.67} & \textbf{70.22} & \textbf{49.00} & \textbf{59.71} & \textbf{65.15} \\
    \midrule
    
    %===== GPT3.5 =====%
    \multirow{8}{*}{\rotatebox[origin=c]{90}{GPT-3.5-turbo}}
      & ICL     & 24.00 & 20.58& 32.69 & 67.62 & 46.00 & 63.72 & 44.62 & 57.30 & 46.15 \\
      & CoT     & 41.79 & 45.45 & 65.66 & 58.83 & 40.54 & 58.60 & 45.94 & 58.70 & 47.06 \\
      & CoT-SC  & 62.00 & 46.33 & 67.31 & 72.83 & 54.74 & 69.03 & 47.28 & 59.90 & 54.62 \\
      & SoT     & 43.42 & 46.92 & 64.25 & 59.80 & 40.83 & 59.04 & 47.68 & 60.91 & 48.21 \\
      & CAD     & — & — & — & 71.79 & 52.66 & 67.45 & 45.05 & 62.66 & 64.53 \\
      & DeCoT   & — & — & — & 70.20 & 53.91 & 68.35 & 46.56 & 63.59 & 64.69 \\
      & CP      & \textbf{84.18} & \textbf{48.36} & \textbf{78.03} & 73.97 & 57.05 & 72.51 & 75.16 & 63.61 & 77.13 \\
      & ACPS    & 81.50 & 48.33 & 74.90 & \textbf{84.45} & \textbf{61.31} & \textbf{75.97} & \textbf{77.01} & \textbf{67.15} & \textbf{79.48} \\
    \bottomrule
    \end{tabular}
    \caption{Performance comparison across seven reasoning datasets. Accuracy (Acc) (\%) is reported for GSM8K, MATH, ComQA, StrQA, and FEVER; Exact Match (EM) and F1 scores (\%) are reported for HotpotQA and MuSiQue. The best results are shown in \textbf{bold}. A dash (–) indicates that the method is not applicable to the dataset, typically because it is designed specifically for knowledge-intensive tasks with external knowledge.}
    \label{tab:main}
\end{table*}

\section{Experiments}
\subsection{Datasets and Evaluation Setup}
Following previous study~\cite{zhang2024causal}, we evaluate our framework across four categories of reasoning tasks to comprehensively assess its performance: math reasoning (GSM8K~\cite{cobbe2021training} and MATH~\cite{hendrycks2021measuring}), commonsense reasoning (CommonsenseQA (ComQA)~\cite{talmor2018commonsenseqa} and StrategyQA (StrQA)~\cite{geva2021did}), multihop reasoning (HotpotQA~\cite{yang2018hotpotqa} and MuSiQue~\cite{trivedi2021musique}), and fact verification (FEVER~\cite{schuster2019towards}). Detailed descriptions of the datasets and the evaluation setup are provided in Appendix~\ref{appendix:dataset_details} and Appendix~\ref{appendix:Evaluation}, respectively.

\subsection{Baseline Methods and Backbone LLMs}

We evaluate our framework against representative baselines, including ICL~\cite{Brown2020GPT3}, CoT~\cite{Wei2022CoT}, CoT-SC~\cite{Wang2023SelfCons}, SoT~\cite{Aytes2025SketchOfThought}, CAD~\cite{Shi2024Trust}, DeCoT~\cite{Wu2024DeCoT}, and CP~\cite{zhang2024causal}. Further details are provided in Appendix~\ref{appendix:baselines}. 

We select three backbone LLMs: Ministral-3B~\cite{MistralAI_Ministral3BInstruct2024}, LLaMA-3.1 8B (LLaMA-3)~\cite{Grattafiori2024Llama3Herd}, and GPT-3.5-turbo~\cite{openai_chatgpt_2022}. These models differ in parameter scale and accessibility (open-source versus closed-source), providing a diverse and balanced foundation for comparison in our evaluation.

\subsection{Main Results}
Table~\ref{tab:main} presents the results of ACPS on three backbone LLMs across seven datasets. ACPS consistently achieves the highest or near-highest scores across all benchmarks. In particular, for context-free tasks that do not rely on external knowledge (e.g., GSM8K, MATH, ComQA), ACPS outperforms existing methods by a large margin. For instance, on GSM8K, it improves over CP by 10.04 points on LLaMA-3, indicating its effectiveness even in complex mathematical reasoning. For contextual tasks that require integration of external knowledge (e.g., HotpotQA, MuSiQue, FEVER), ACPS continues to lead, surpassing CAD and DeCoT. On HotpotQA with GPT-3.5-turbo, ACPS obtains 61.31 EM and 75.97 F1, outperforming CP (57.05 EM, 72.51 F1) and DeCoT (53.91 EM, 68.35 F1). Similar gains are observed on MuSiQue and FEVER. These results highlight ACPS’s ability to adapt to both types of reasoning by selecting an appropriate intervention and leveraging efficient SoT-based reasoning. Its consistent superiority across datasets and model scales demonstrates the generalisability of the proposed framework.

\subsection{Efficiency Analysis}

To assess the efficiency of various prompting methods, we conduct a comprehensive analysis across multiple reasoning tasks. Specifically, we measure (i) the average number of reasoning steps taken, (ii) the average number of tokens consumed (in Appendix~\ref{Consumption}), and (iii) the accuracy-efficiency trade-off under token budgets (in Appendix~\ref{Accuracy-Efficiency}). Figure~\ref{fig:avg_step_ca} presents the average number of reasoning steps taken on seven datasets. CoT and CoT-SC often produce lengthy reasoning chains, with more than six steps on datasets such as HotpotQA. In contrast, ACPS consistently generates shorter reasoning traces, typically fewer than three steps, while maintaining strong performance. This demonstrates the superior token efficiency of {ACPS} compared to more verbose causality-based prompting methods like {DeCoT} and {CP}. Further efficiency analyses are provided in Appendix~\ref{appendix:eff_analysis}.

\begin{table*}[t]
  \small
  \setlength{\tabcolsep}{8pt}
  \renewcommand{\arraystretch}{0.8}
  \centering
  \begin{tabular}{l
                  |cc   % GSM8K Acc
                  |cc   % CommonsenseQA Acc
                  |cc   % HotpotQA EM/F1
                  |cc   % MuSiQue EM/F1
                  |c}   % FEVER Acc
    \toprule
    & {GSM8K} & {MATH}  & {ComQA} & {StrQA} 
    & \multicolumn{2}{c}{{HotpotQA}}
    & \multicolumn{2}{c|}{{MuSiQue}}
    & {FEVER}\\
      \cmidrule(lr){2-2}
      \cmidrule(lr){3-3}
      \cmidrule(lr){4-4}
      \cmidrule(lr){5-5}
      \cmidrule(lr){6-7}
      \cmidrule(lr){8-9}
      \cmidrule(lr){10-10}
    & {Acc}~$\uparrow$ & {Acc}~$\uparrow$ & {Acc}~$\uparrow$ & {Acc}~$\uparrow$ 
    & {EM}~$\uparrow$ & {F1}~$\uparrow$
    & {EM}~$\uparrow$ & {F1}~$\uparrow$
    & {Acc}~$\uparrow$ \\
    \midrule
    ACPS & \textbf{81.50} & 48.33 & 74.90 & \textbf{84.45} & \textbf{61.31} & \textbf{75.97} & \textbf{77.01} &\textbf{67.15} & 79.48 \\
    ~~w/o SoT & 80.75 & \textbf{52.95} & \textbf{77.11} & 83.17 & 60.15 & 75.26 & 75.15 & 66.45 & \textbf{83.20} \\
    ~~NWGM-Rev & 81.47 & 48.05 & 74.21 & 84.05 & 60.67 & 74.66 & 76.89 & 66.67 & 78.89 \\
    ~~NWGM-Ran  & 81.25 & 46.67 & 73.89 & 82.63 & 59.21 & 73.41 & 75.33 & 64.01 & 74.40 \\
    ~~w/o K-means   & 80.95 & 44.63 & 72.25 & 81.50 & 57.32 & 68.92 & 74.13 & 62.30 & 73.15 \\
    ~~w/o Weight     & 78.57 & 41.75 & 71.43 & 77.80 & 55.87 & 66.97 & 62.45 & 60.71 & 71.41 \\
    \bottomrule
  \end{tabular}
  \caption{Ablation study results on seven datasets using GPT-3.5-turbo. The best results are shown in \textbf{bold}.}
  \label{tab:ablation}
\end{table*}

\begin{table}[t]
\small
\renewcommand{\arraystretch}{0.8}
\setlength\tabcolsep{3.5pt}
\centering
\begin{tabular}{l|cc|cc|cc}
    \toprule
     & 
    \multicolumn{2}{c|}{{HotpotQA}} & 
    \multicolumn{2}{c|}{{HotpotQA-Inj}} & 
    \multicolumn{2}{c}{{HotpotQA-Shuf}} \\ \cmidrule{2-7} 
     {Method}& {EM}~$\uparrow$ & {F1}~$\uparrow$ & {EM}~$\uparrow$ & {F1}~$\uparrow$ & {EM}~$\uparrow$ & {F1}~$\uparrow$ \\
    \midrule
    ICL      & 46.00 & 63.72 & 31.05 & 42.30 & 52.89 & 66.99 \\
    CoT      & 40.54 & 58.60 & 31.35 & 42.28 & 52.77 & 67.45 \\
    CoT-SC   & 54.74 & 69.03 & 28.57 & 40.18 & 51.97 & 66.31 \\
    SoT      & 40.83 & 59.04 & 32.01 & 43.85 & 53.15 & 66.55 \\
    CAD      & 52.66 & 67.45 & 29.11 & 42.40 & 52.41 & 66.59 \\
    DeCoT    & 53.91 & 68.35 & 29.47 & 26.89 & 51.85 & 67.43 \\
    CP       & 57.05 & 72.51 & 34.55 & 47.97 & 51.27 & 67.87 \\
    ACPS     & \textbf{61.31} & \textbf{75.97} & \textbf{37.68} & \textbf{48.78} & \textbf{60.20} & \textbf{75.84} \\
    \bottomrule
\end{tabular}
\caption{Robustness results on the HotpotQA dataset using GPT-3.5-turbo. The best results are shown in \textbf{bold}.}

\label{tab:robustness}
\end{table}

\subsection{Robustness Study}
To assess the robustness of our framework under noisy and disordered scenarios, we conduct experiments on the HotpotQA dataset with two types of data perturbation. In HotpotQA-Inj, one evidence item per record is replaced with unrelated content, introducing semantic noise. In HotpotQA-Shuf, the order of evidence items is shuffled to disrupt contextual flow. These settings evaluate the model's ability to reason under degraded input conditions. As shown in Table~\ref{tab:robustness}, ACPS consistently achieves the best performance across the original, injected, and shuffled versions of HotpotQA. In the injected setting, ACPS achieves 3.1\% higher EM and 0.8\% higher F1 than CP. Under the shuffled setting, ACPS significantly outperforms CP by 8.9\% in EM and 8.0\% in F1. These results demonstrate that ACPS is robust to input perturbations and effectively handles noisy or disordered inputs through causal SoT-based reasoning.

\begin{figure}[t]
    \centering
    \includegraphics[width=0.48\textwidth]{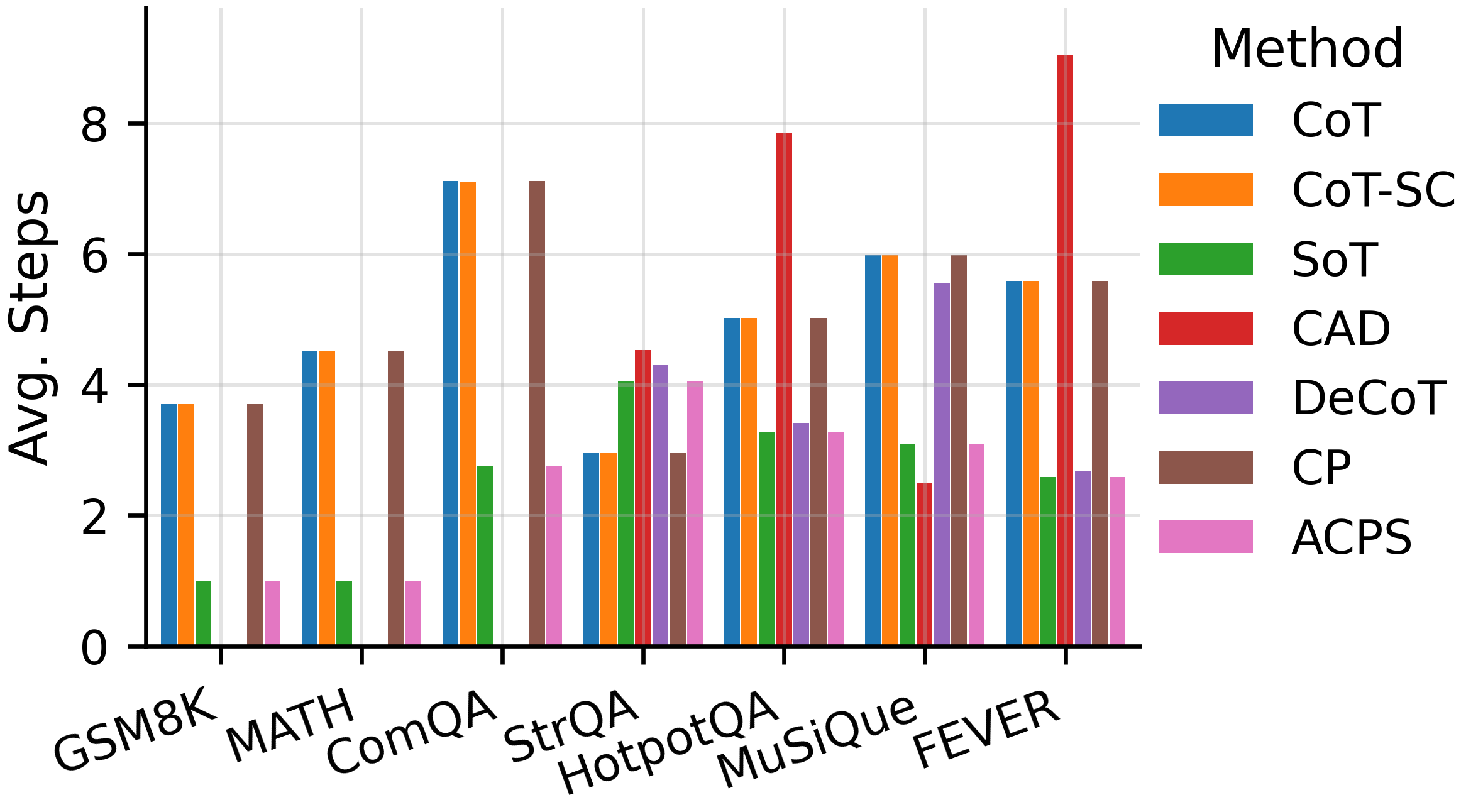}
    \caption{Comparison of the average number of reasoning steps across all datasets for different prompting methods.}
    \label{fig:avg_step_ca}
\end{figure}

\subsection{Ablation Study}
We conduct an ablation study across seven reasoning datasets using GPT-3.5-turbo to evaluate the contribution of key components in ACPS. As shown in Table~\ref{tab:ablation}, we compare the full model with several variants: {w/o SoT} (replacing SoT with CoT), {NWGM-Rev} (reversing the order of in-context examples), {NWGM-Ran} (random example selection), {w/o K-means} (removing clustering), and {w/o Weight} (omitting causal-effect-based ranking). While the w/o SoT variant achieves the best results on MATH, ComQA, and FEVER, SoT offers consistent advantages on most datasets and remains competitive overall, highlighting its value as a concise yet effective reasoning paradigm. Other ablations cause notable drops, with the largest from {w/o Weight} (e.g., MuSiQue F1: 67.15~$\rightarrow$~60.71), confirming the importance of causal-effect-based selection. Removing K-means also reduces performance (e.g., HotpotQA F1: 75.97~$\rightarrow$~68.92), showing the benefit of reasoning diversity. Moreover, {NWGM-Ran} underperforms {NWGM-Rev}, indicating that example relevance matters more than order. Overall, these results demonstrate that both similarity-guided prompt construction and causal-effect-aware selection are essential for robust reasoning in ACPS.

% We conduct an ablation study across seven reasoning datasets using GPT-3.5-turbo to evaluate the contribution of key components in ACPS. As shown in Table~\ref{tab:ablation}, we compare the full model with several variants: {w/o SoT} (replacing SoT with CoT), {NWGM-Rev} (reversing the order of in-context examples), {NWGM-Ran} (random example selection), {w/o K-means} (removing clustering for diverse reasoning traces), and {w/o Weight} (omitting causal-effect-based ranking). While the w/o SoT variant achieves the best performance on MATH, ComQA, and FEVER, SoT provides consistent advantages on the majority of datasets and remains competitive overall, underscoring its value as a concise yet effective reasoning paradigm within ACPS. Other ablations lead to notable performance drops, with the most substantial degradation observed in {w/o Weight} (e.g., MuSiQue F1: 67.15~$\rightarrow$~60.71), confirming the importance of causal-effect-based output selection. Removing K-means also results in consistent declines (e.g., HotpotQA F1: 75.97~$\rightarrow$~68.92), highlighting the role of reasoning diversity. Moreover, {NWGM-Ran} underperforms {NWGM-Rev}, indicating that the relevance of examples outweighs their order. Collectively, these results demonstrate that both similarity-guided prompt construction and causal-effect-aware selection are critical for achieving robust and accurate reasoning in ACPS.

\subsection{Hyper-parameter Study}
We conduct a hyper-parameter study on the number of generated SoTs (\(M\)) and clusters (\(K\)). We observe that larger values generally improve performance but incur higher computational cost. The complete results are provided in Appendix~\ref{appendix:hyper-study}.

\subsection{Additional Details}
Due to page limitations, further implementation details, including the core component setup and demonstration step, are provided in Appendix~\ref{Implementation}. The prompting template is given in Appendix~\ref{appendix:promtp_templates}, and case studies are presented in Appendix~\ref{Case}.

\section{Related Work}
Mitigating biases in LLMs increasingly relies on causal inference frameworks~\cite{Pearl2009Causality, pearl2016causal,XuCLLLW23}. With strong theoretical guarantees, various methods have been developed to estimate causal effects even in the presence of unobserved confounders~\cite{Cheng2024Disentangled, Cheng2024LongitudinalIV,DuLC0GC025}. As a result, LLM reasoning has been increasingly framed within SCMs to reduce spurious correlations. For example, DeCoT applies front-door adjustment using CoT as a mediator and external knowledge as an instrumental variable, improving reasoning performance but limiting generality due to its reliance on external inputs~\cite{Wu2024DeCoT}. Similarly, Causal Prompting employs CoT-based front-door adjustment enhanced by contrastive learning; however, its causal guarantees do not generalise to all task types~\cite{zhang2024causal}. See Appendix~\ref{related} for more related work.

Our framework differs from prior methods by unifying the standard and conditional front-door adjustments within a single framework, addressing both context-free and context-dependent reasoning tasks. We further incorporate SoT to improve inference efficiency. This design enhances scalability and generality, offering an effective and principled solution for mitigating bias in LLMs.

\section{Conclusion}
In this paper, we present ACPS, a novel prompting framework that integrates standard and conditional front-door adjustments with efficient Sketch-of-Thought reasoning. By adaptively selecting the appropriate intervention based on task characteristics, ACPS mitigates internal biases in large language models while substantially reducing token usage and inference cost. Extensive experiments demonstrate that ACPS consistently outperforms existing prompting baselines in accuracy, efficiency, and robustness.

% At the end of the paper:
% after the dicussion/conclusion section
% before the references
% it does not count toward the page limit

\section{Limitations}
\label{limitation}

Although our results demonstrate the effectiveness of our framework, several aspects warrant further exploration. Expanding the evaluation to larger-scale test sets and more powerful backbone models could provide deeper insights into the generalizability and scalability of the framework. In addition, while we vary the temperature to encourage diversity in SoTs, certain edge-case generations remain constrained by the API’s safety policy; future work may consider strategies to address such cases while remaining compliant with safety requirements. Finally, further validation on broader datasets and in real-world scenarios would help strengthen the evidence of robustness and practical applicability.

\bibliography{custom}

\appendix
% \setcounter{secnumdepth}{2}

% \renewcommand{\thesection}{\Alph{section}} 
% \renewcommand{\thesubsection}{\thesection.\arabic{subsection}}

% \twocolumn[
% \begin{center}
%     \LARGE \textbf{Appendix for ``Debiasing Large Language Models via Adaptive Causal Prompting with Sketch-of-Thought"}
%     \vspace{1cm}
% \end{center}
% ]

\section{Discussion}

\subsection{Why a single unobserved confounder $U$?}

Although real-world reasoning may involve multiple unobserved confounders, our assumption of a single unobserved confounder in the SCM is consistent with prior work and facilitates tractable causal analysis via the front-door criterion. Empirically, our experiments show that LLMs often initiate reasoning correctly but tend to fail at the final step due to internal biases. This pattern indicates the presence of a dominant confounder that substantially influences the direct relationship between the query and the answer, thereby supporting the validity of our simplified SCM. Moreover, the ACPS framework addresses potential biases introduced by reasoning traces by generating diverse SoTs through clustering and applying the NWGM algorithm for prompt selection. Consequently, the single unobserved  confounder assumption proves to be both practically robust and computationally efficient. Future work will consider more complex causal structures to model additional confounding factors more comprehensively.

\subsection{Why not fine-tune the encoder?}

\begin{figure}[t]
    \centering
    \begin{subfigure}[b]{0.45\textwidth}
        \centering
        \includegraphics[width=\textwidth]{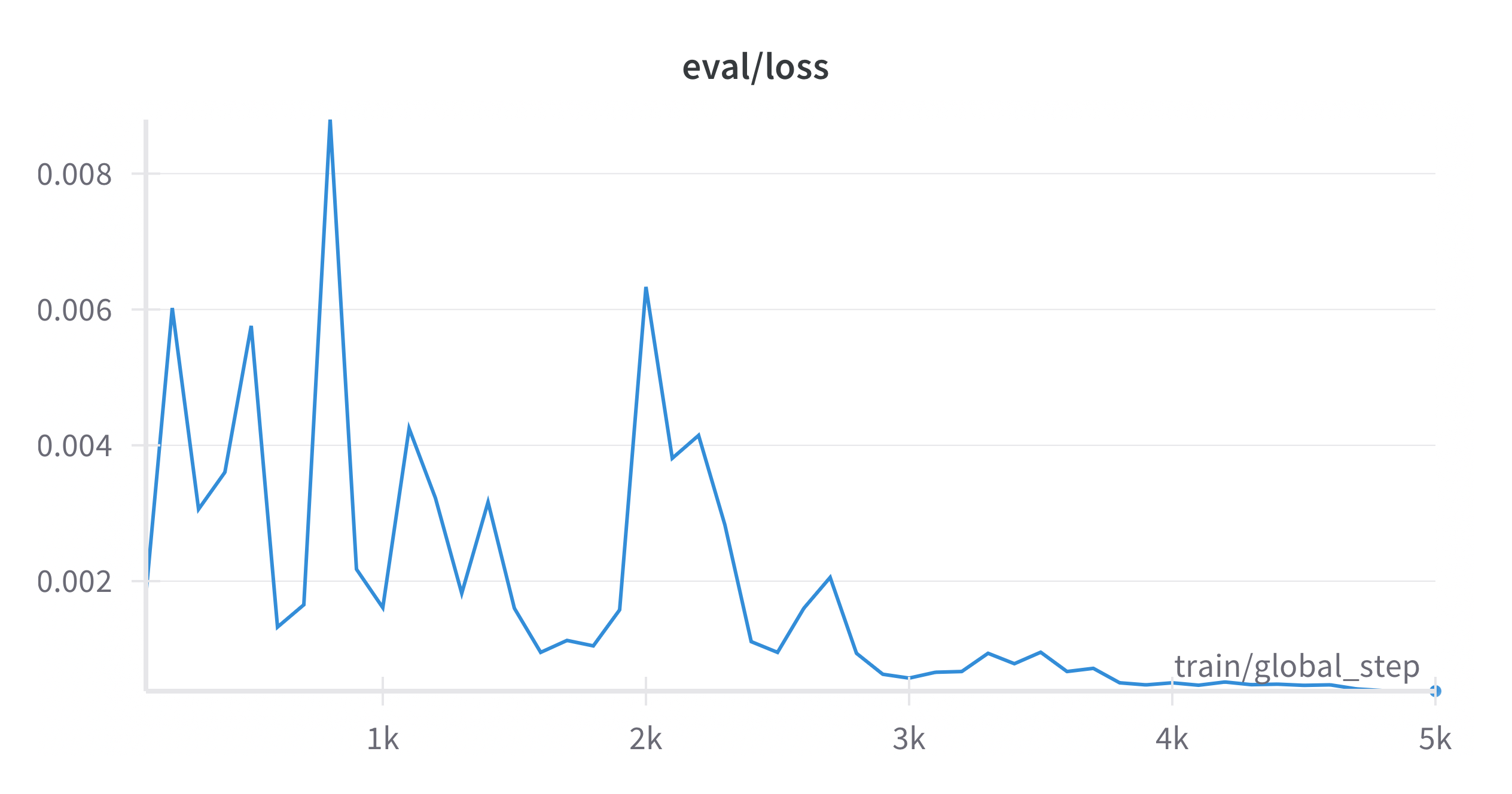}
        \caption{}
        \label{fig:eval_loss}
    \end{subfigure}
    \hfill
    \begin{subfigure}[b]{0.45\textwidth}
        \centering
        \includegraphics[width=\textwidth]{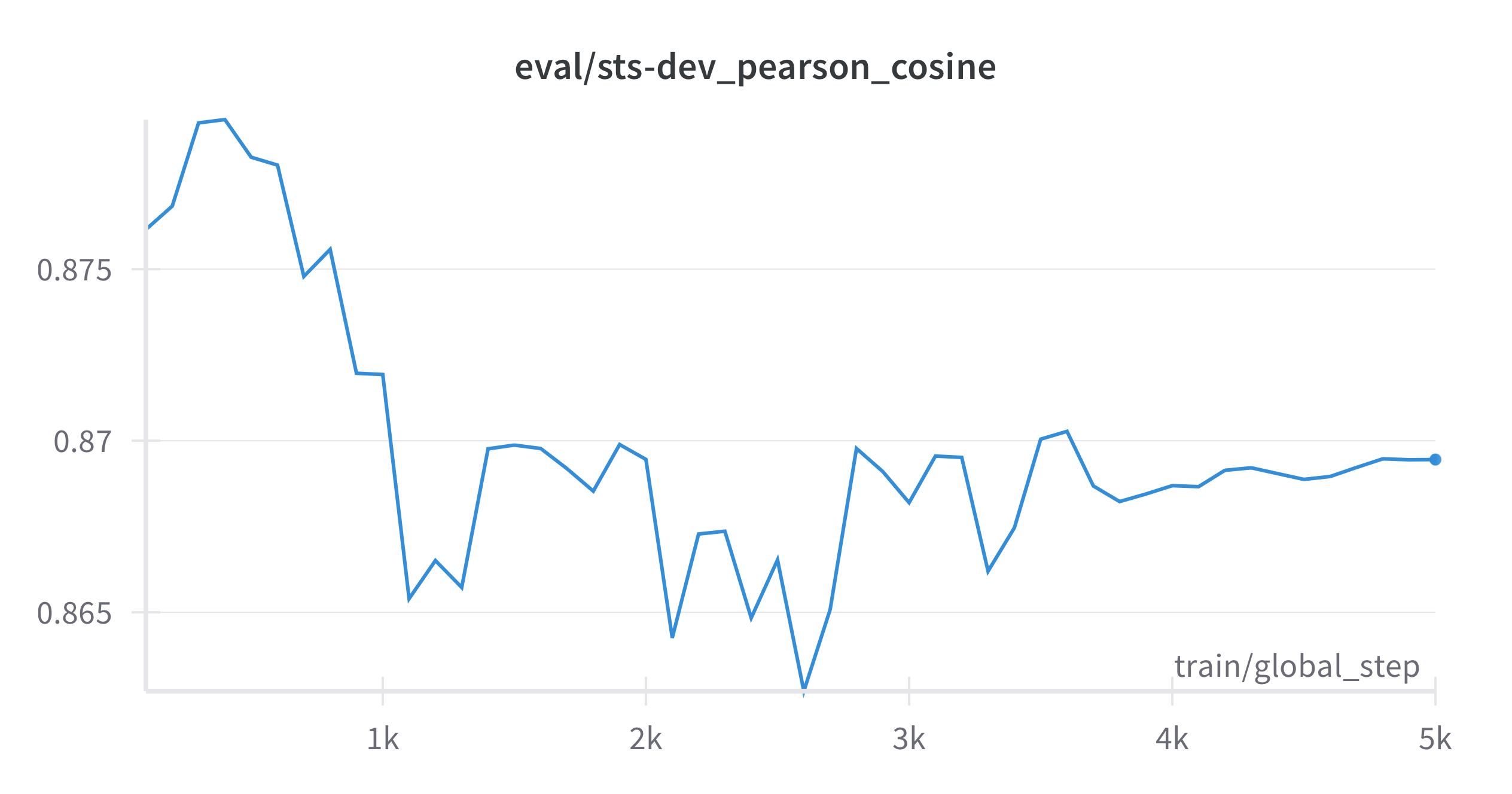}
        \caption{}
        \label{fig:sts_dev_pearson_cosine}
    \end{subfigure}
    \caption{Trends of (a) evaluation loss and (b) \texttt{sts-dev\_pearson\_cosine} during encoder fine-tuning on GSM8K.}
    \label{fig:unstable_curves}
\end{figure}

We fine-tune the encoder using the \texttt{SentenceTransformerTrainer}~\cite{reimers-2019-sentence-bert} on GSM8K (4,096 examples) with 20 epochs, a batch size of 16, and a learning rate of $1 \times 10^{-4}$. However, the training shows unstable evaluation loss and highly variable correlation metrics (\texttt{sts-dev\_pearson\_cosine}) (see Figure~\ref{fig:unstable_curves}). In addition, a persistent gap between training and evaluation loss indicates rapid overfitting, suggesting that the dataset is too small to support reliable fine-tuning. To avoid these issues, we use the pre-trained \texttt{Sentence-BERT} encoder without additional fine-tuning, which provides more stable embeddings and better generalisation across tasks.

\subsection{What happens if classification fails?}

When the classification engine fails to produce a reliable output, we employ a default template grounded in commonsense knowledge (CS). This template is applicable to both contextualised and uncontextualised tasks, thereby ensuring robust and consistent performance across all task types.

\section{Detailed Derivations}
\label{appendix:derivation}

\begin{theorem}[Rules of $do$-Calculus~\citep{Pearl2009Causality}]
    \label{do}
    Let $\mathcal{G}$ be the DAG associated with a structural causal model, and let $P(\cdot)$ denote the probability distribution induced by that model. For any disjoint subsets of variables $Q, A, Z$, and $W$, the following rules hold:
    \begin{itemize}[leftmargin=0.4cm]
        \item Rule 1. (Insertion/deletion of observations):
        \[
        P(A \mid do(A), Z, W) = P(A \mid do(Q), W)\]
        \[
        \quad\quad \text{if } (A \indep Z \mid A, W) \text{ in } \mathcal{G}_{\overline{Q}}.
        \]
        \item Rule 2. (Action/observation exchange):
        \[
        P(A \mid do(Q), do(Z), W) = P(A \mid do(Q), Z, W),\]
        \[
        \quad\quad \text{if } (Y \indep Z \mid Q, W) \text{ in } \mathcal{G}_{\overline{Q}\underline{Z}}.
        \]
        \item Rule 3. (Insertion/deletion of actions):
        \[
        P(A \mid do(Q), do(Z), W) = P(A \mid do(Q), W),
        \]
        \[
        \quad\quad \text{if } (A \indep Z \mid Q, W) \text{ in } \mathcal{G}_{\overline{Q}, \overline{Z(W)}}.
        \]
        where $Z(W)$ is the set of nodes in $Z$ that are not ancestors of any node in $W$ in $\mathcal{G}_{\overline{Q}}$.
    \end{itemize}
\end{theorem}

As shown in Figure~\ref{fig2:c}, \(R\) meets all the requirements of the conditional front-door criterion for the pair \((Q,A)\), with the external knowledge variable \(E\) acting as the conditioning set \(W\). Thus, \(R\) serves as a valid conditional front-door adjustment variable to identify the causal effect of \(Q\) and \(A\). We now apply Theorem 2 to to derive \( P(A \mid do(Q)) \), with the derivation process detailed below:
\scalebox{0.65}{$
\begin{aligned}
\centering
    &P(A \mid do(Q)) = \sum_{r}P(r\mid do(Q)) P(A\mid r,do(Q)) \nonumber\\
    &~= \sum_{r}P(r\mid do(Q))\sum_{e} P(A \mid \mathrm{do}(Q),r,e) P(e\mid \mathrm{do}(Q),r) \nonumber\\
    &~= \sum_{r}P(r\mid do(Q))\sum_{e} P(A \mid \mathrm{do}(Q),\mathrm{do}(r),e) P(e\mid \mathrm{do}(Q),r), \nonumber \\
     &\quad \text{since }  (A \indep R\mid Q, E) \text{ in } \mathcal{G}_{\overline{Q}\underline{R}}~\text{(Rule 2 in Theorem~\ref{do})} \nonumber\\
    &~= \sum_{r}P(r\mid do(Q))\sum_{e} P(A \mid \mathrm{do}(r),e) P(e\mid \mathrm{do}(Q),r), \nonumber\\
    &\quad \text{since }  (A \indep Q \mid R, E) \text{ in } \mathcal{G}_{\overline{R}\overline{Q(E)}}~\text{(Rule 3 in Theorem~\ref{do})} \nonumber\\
    &~= \sum_{r}P(r\mid do(Q))\sum_{e,~q} P(A \mid \mathrm{do}(r),q,e) P(q \mid \mathrm{do}(r),e) P(e\mid \mathrm{do}(Q),r), \nonumber\\
    &~= \sum_{r}P(r\mid do(Q))\sum_{e,~q} P(A \mid r,q,e) P(q \mid \mathrm{do}(r),e) P(e\mid \mathrm{do}(Q),r), \nonumber \\
    &\quad \text{since }  (A \indep R \mid Q, E) \text{ in } \mathcal{G}_{\underline{R}}~\text{(Rule 2 in Theorem~\ref{do})} \nonumber\\
    &~= \sum_{r}P(r\mid do(Q))\sum_{e,~q} P(A \mid r,q,e) P(q \mid e) P(e\mid \mathrm{do}(Q),r), \nonumber \\
    &\quad\text{since }  (Q \indep R \mid E) \text{ in } \mathcal{G}_{\overline{R(E)}}~\text{(Rule 3 in Theorem~\ref{do})}\nonumber
\end{aligned}
$}
\scalebox{0.65}{$
\begin{aligned}
\centering
    &~= \sum_{r}P(r\mid do(Q))\sum_{e,~q} P(A \mid r,q,e) P(q \mid e) \frac{P(e, r \mid do(Q))}{P(r \mid \mathrm{do}(Q))}, \nonumber \\
     &\quad \text{since the chain rule of conditional probability}\nonumber\\
    &~= \sum_{r}P(r\mid do(Q))\sum_{e,~q} P(A \mid r,q,e) P(q \mid e) \frac{P(r \mid Q, e) p(e)}{P(r \mid \mathrm{do}(Q))} \nonumber \\
    &~= {\sum_{r,~e} P(r \mid Q, e)} {\sum_{q,~e} P(A \mid r,q,e) P(q \mid e) P(e)} \nonumber
\end{aligned}
$}

In our framework, \(Q\) represents the fixed input query during reasoning. Since no intervention is applied to \(Q\), it is considered a constant instead of a random variable. Therefore, the term \( P(q \mid e) \) and the summation over \(q\) can be removed, simplifying the causal effect expression as follows:

{\small
\begin{equation}
    \centering
    P(A \mid do(Q)) = \sum_{r,~q,~e}  \underbrace{P(r \mid Q, e)}_{\text{\ding{172}}} \underbrace{P(A \mid r,q,e)}_{\text{\ding{173}}}\underbrace{P(e)}_{\text{\ding{174}}}\nonumber
\end{equation}}

\section{ACPS Algorithm}
\label{appendix:Algorithm}
We describe the ACPS procedure in detail in Algorithm~\ref{alg:tacp}.

\begin{algorithm}[t]
\caption{ACPS}
\label{alg:tacp}
\begin{algorithmic}[1]
\Statex \textbf{Input:} Query \( Q \), External knowledge \( E \), Encoder, D, d, LLM
\Statex \textbf{Parameters:} 
$M$ (number of initial SoTs), 
$K$ (number of clusters),
\State $prompt \gets [d_{1},...d_{n}, \{Q,E\}] $
\State $R_{\text{init}} = [r_1, r_2, \dots, r_m] \gets \text{LLM}(prompt)$
\State $\tilde{R}_{\text{init}} \gets \text{Encoder}(R_{\text{init}})$
\State $\tilde{R} = [\tilde{r}_1, \tilde{r}_2, \dots, \tilde{r}_k] \gets \text{K-means}(\tilde{R}_{\text{init}})$
% \State Compute \( P(E) \) using Eq.~\ref{1}
\For{$n = 1$ to $N$}
    \State Compute \( P(r_k \mid Q, e) \) using Eq.~\ref{equation_r_k} and ~\ref{222}
    \State Compute \( P(A \mid r_k, Q, e) \) using Eq.~\ref{re_rank}, ~\ref{intervention_r_k} and ~\ref{333}
\EndFor
\State Compute \( P(A \mid do(Q)) \) using Eq.~\ref{666} 
\State \Return \( \displaystyle\arg\max_{A} P(A = a \mid do(Q)) \)
\end{algorithmic}
\end{algorithm}

\section{Experimental Details}

\subsection{Dataset Details}
\label{appendix:dataset_details}
\begin{itemize}
    \item \textbf{GSM8K}~\cite{cobbe2021training} comprises 8.5K grade-school mathematics word problems requiring 2–8 arithmetic steps. 
    
    \item \textbf{MATH}~\cite{hendrycks2021measuring} contains 12.5K competition-level mathematics problems with detailed step-by-step solutions. 

    \item \textbf{CommonsenseQA}~\cite{talmor2018commonsenseqa} includes 12,247 multiple-choice questions grounded in ConceptNet relations, designed to probe background world knowledge. 

    \item \textbf{StrategyQA}~\cite{geva2021did} is a yes/no question answering benchmark comprising 2,780 questions that require implicit multi-step reasoning. Each question is annotated with a decomposition and supporting Wikipedia paragraphs.

    \item \textbf{HotpotQA}~\cite{yang2018hotpotqa} consists of 113K Wikipedia-based question–answer pairs that require multi-hop reasoning across documents. 

    \item \textbf{MuSiQue}~\cite{trivedi2021musique} contains approximately 25K two- to four-hop questions constructed by composing single-hop questions from existing datasets.

    \item \textbf{Symmetric FEVER}~\cite{schuster2019towards} is a diagnostically enhanced evaluation set derived from the original FEVER benchmark, comprising 1,420 counterfactual claim–evidence pairs. Each instance contains one supporting and one refuting Wikipedia sentence, labelled as \textsc{SUPPORT} or \textsc{REFUTE}, specifically constructed to expose and mitigate claim-only biases in fact-verification systems. For brevity, we refer to Symmetric FEVER as “FEVER” throughout the remainder of this paper.
\end{itemize}

All datasets used in this work (GSM8K, MATH, CommonsenseQA, StrategyQA, HotpotQA, MuSiQue, and FEVER) are publicly available under their respective research licenses and are used strictly for research purposes, consistent with their intended use. These benchmarks do not contain personally identifying information. While some datasets may include open-domain text with potentially sensitive content, they have been widely adopted in prior work and released with appropriate safeguards. We did not collect new data, and no additional anonymization was necessary.

\subsection{Evaluation Setup}
\label{appendix:Evaluation}
Consistent with prior work~\cite{Lyu2023Faithful, zhang2024causal}, we evaluate the performance of ACPS across different reasoning paradigms. For Chunked Symbolism tasks, we use label classification accuracy (Acc), which is appropriate for mathematical reasoning involving numerical and symbolic operations. For Conceptual Chaining tasks, including CommonsenseQA and StrategyQA, we also use accuracy, given the nature of these tasks, which require connecting ideas in logical sequences. For Multihop Reasoning tasks, we adopt Exact Match (EM) and F1 scores, as these metrics are better suited for problems that require multi-step reasoning across multiple pieces of information. In line with~\citet{Aytes2025SketchOfThought}, we extract the answer text span enclosed within the \texttt{\textbackslash boxed\{\}} keyword when evaluating span-based reasoning tasks. The dataset setup details are provided in Table~\ref{tab:dataset_details}.

\begin{table}[t]
\centering
    \small{\begin{tabular}{ccc}
    \toprule
    \textbf{Dataset} & \textbf{Measure / Eval Type} \\
    \midrule
    GSM8K   & Accuracy / numerical \\
    MATH     & Accuracy / open \\
    \midrule
    CommonsenseQA  & Accuracy / multiple\_choice \\
    StrategyQA     & Accuracy / yes/no \\
    \midrule
    HotpotQA  & F1 \& EM / open \\
    MuSiQue  & F1 \& EM / open \\
    \midrule
    FEVER & Accuracy / supports/refutes \\
    \bottomrule
    \end{tabular}}
\caption{Details of dataset setup for experiments.}
\label{tab:dataset_details}
\end{table}

\subsection{Baseline Methods}
\label{appendix:baselines}
We briefly describe the baseline methods considered in our evaluation:
\begin{itemize}
    \item \textbf{In-Context Learning (ICL)}~\cite{Brown2020GPT3}: Uses input–output demonstrations without explicit reasoning steps.
    \item \textbf{Chain-of-Thought (CoT)}~\cite{Wei2022CoT}: Incorporates intermediate reasoning steps to support logical inference.
    \item \textbf{CoT with Self-Consistency (CoT-SC)}~\cite{Wang2023SelfCons}: Samples multiple reasoning paths and selects the majority answer.
    \item \textbf{Sketch-of-Thought (SoT)}~\cite{Aytes2025SketchOfThought}: Produces concise reasoning sketches that capture essential logic while reducing token usage.
    \item \textbf{Context-Aware Decoding (CAD)}~\cite{Shi2024Trust}: Enhances reliability by comparing model outputs generated with and without additional context.
    \item \textbf{Debiasing CoT (DeCoT)}~\cite{Wu2024DeCoT}: Mitigates bias via front-door adjustment with external knowledge.
    \item \textbf{Causal Prompting (CP)}~\cite{zhang2024causal}: Estimates causal effects between prompts and answers using front-door adjustment.
\end{itemize}

\section{Implementation Details}
\label{Implementation}
\subsection{LLM Setup} 
We design a custom asynchronous client for the Microsoft Azure serverless inference API to support concurrent requests during experimentation.

\subsection{Classification Engine Setup} 
In line with~\cite{Aytes2025SketchOfThought}, we employ the router model as our classification engine. This model assigns queries to distinct reasoning paradigms and is applied without further training or modification, maintaining consistency with established reasoning frameworks.

\subsection{Encoder Setup} 
We use Sentence-BERT~\cite{reimers-2019-sentence-bert}, a pre-trained language model, as the encoder for computing sentence embeddings. The embeddings are then used for similarity measurement, SoT clustering, and in-context demonstration selection within the NWGM algorithm.

\subsection{Demonstration Construction}

We standardise the demonstration construction process across all datasets by designating the answer column (i.e., the ground truth) as the reference label. Rather than relying on gold rationales or manual annotations, we automatically generate both correct and incorrect SoTs by sampling model outputs under different temperature settings. Higher temperatures increase diversity and creativity, yielding a mix of accurate and flawed reasoning paths. For the demonstration set, we randomly select questions associated with both correct and incorrect SoTs to capture diverse reasoning patterns. To rigorously evaluate the debiasing effect, demonstrations are constructed only from the original dataset, while evaluation is performed on both the original and adversarial datasets. This design ensures a consistent and scalable demonstration pipeline applicable across multiple tasks.

\subsection{Demonstration Selection}

We select the most relevant demonstrations for each query based on embedding similarity and concatenate them into the prompt. For each dataset, a dedicated demonstration set is constructed. By default, we use two demonstrations per task (\(l=2\) in Eq.~\ref{intervention_r_k}), ensuring a consistent few-shot prompting configuration across datasets. This design keeps the selection and integration process uniform and reproducible across all tasks in our study.

\subsection{SoT Generation and Answer Selection} 
To minimise computational overhead, we pre-generate all SoTs for each query before embedding computation. We sample SoTs with temperatures ranging from $0.0$ to $2.0$ in increments of $0.25$, while fixing top\_p at $0.9$ to encourage diversity. This yields $M=9$ SoTs per query, which are clustered into $K=4$ groups using K-Means. For each cluster centroid, we generate $S=3$ answers via prompts refined through our causal intervention procedure. The resulting $K \times S = 12$ answers are then aggregated by causality-weighted voting to obtain the final prediction.

\section{Experimental Results}
\subsection{Efficient Analysis}
\label{appendix:eff_analysis}

\subsubsection{Efficiency Comparison of CoT and SoT}

\begin{figure}[t]
    \centering
    \includegraphics[width=0.45\textwidth]{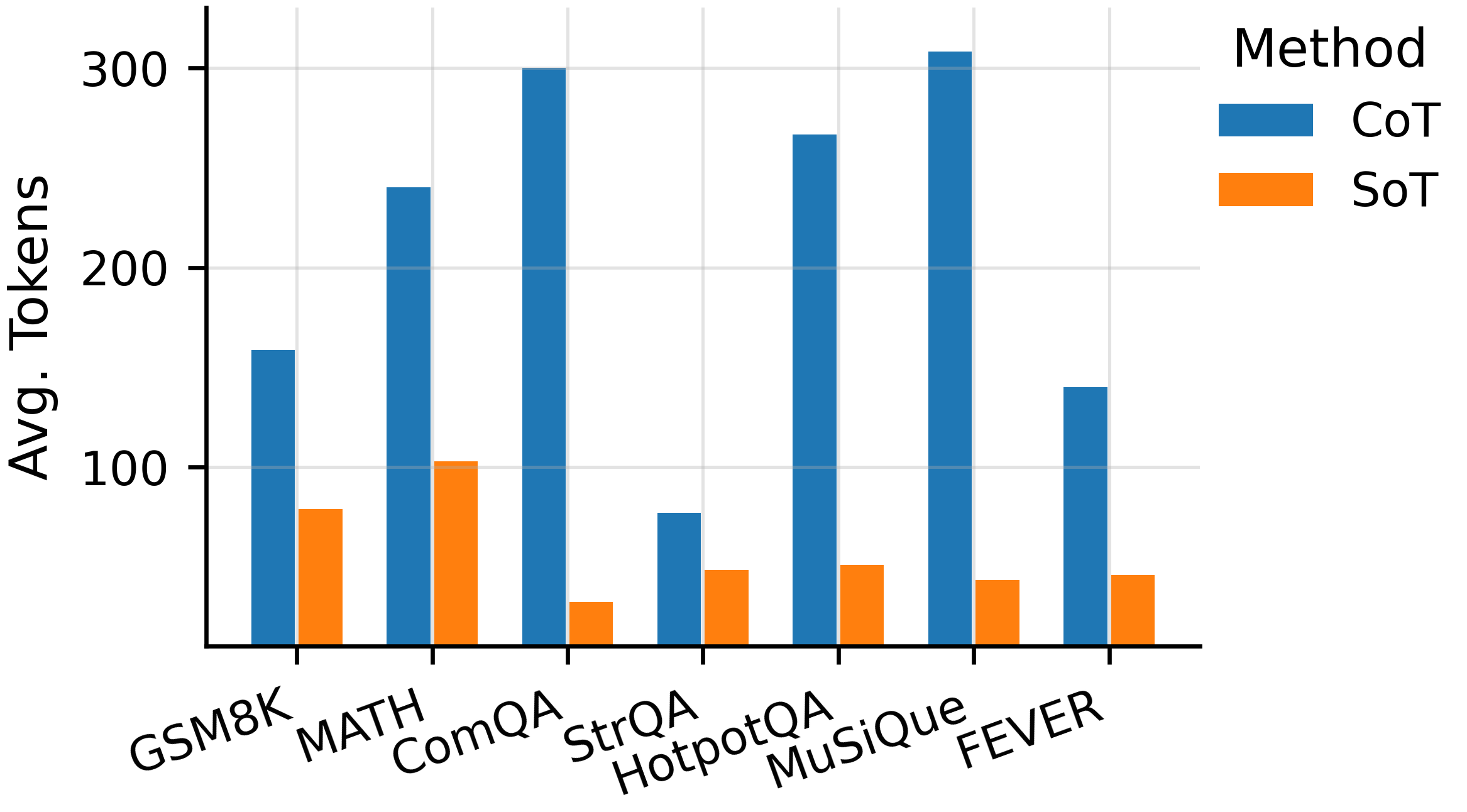}
    \caption{Comparison of average tokens consumed between CoT and SoT.}
    \label{fig:sot_cot_avg_token}
\end{figure}

\begin{figure}[t]
    \centering
    \includegraphics[width=0.45\textwidth]{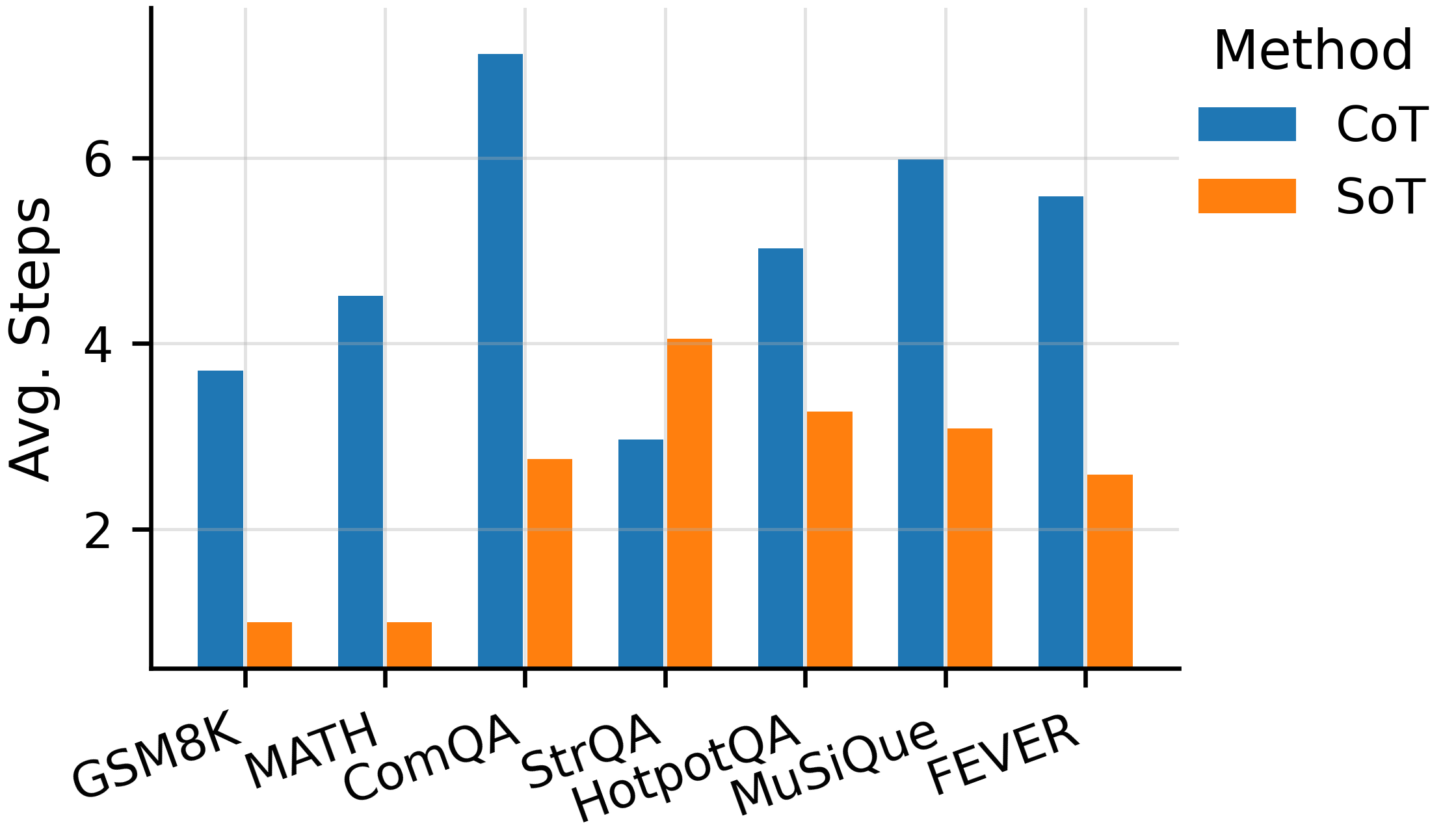}
    \caption{Comparison of average reasoning steps between CoT and SoT.}
    \label{fig:sot_cot_avg_step}
\end{figure}

We evaluate the efficiency of CoT and SoT by comparing the number of tokens and reasoning steps on identical questions across tasks. Figures~\ref{fig:sot_cot_avg_token} and~\ref{fig:sot_cot_avg_step} show that SoT achieves comparable performance with fewer tokens and shorter reasoning paths, highlighting its superior efficiency.

\subsubsection{Token Consumption Analysis}
\label{Consumption}
\begin{figure}[t]
    \centering
    \includegraphics[width=0.45\textwidth]{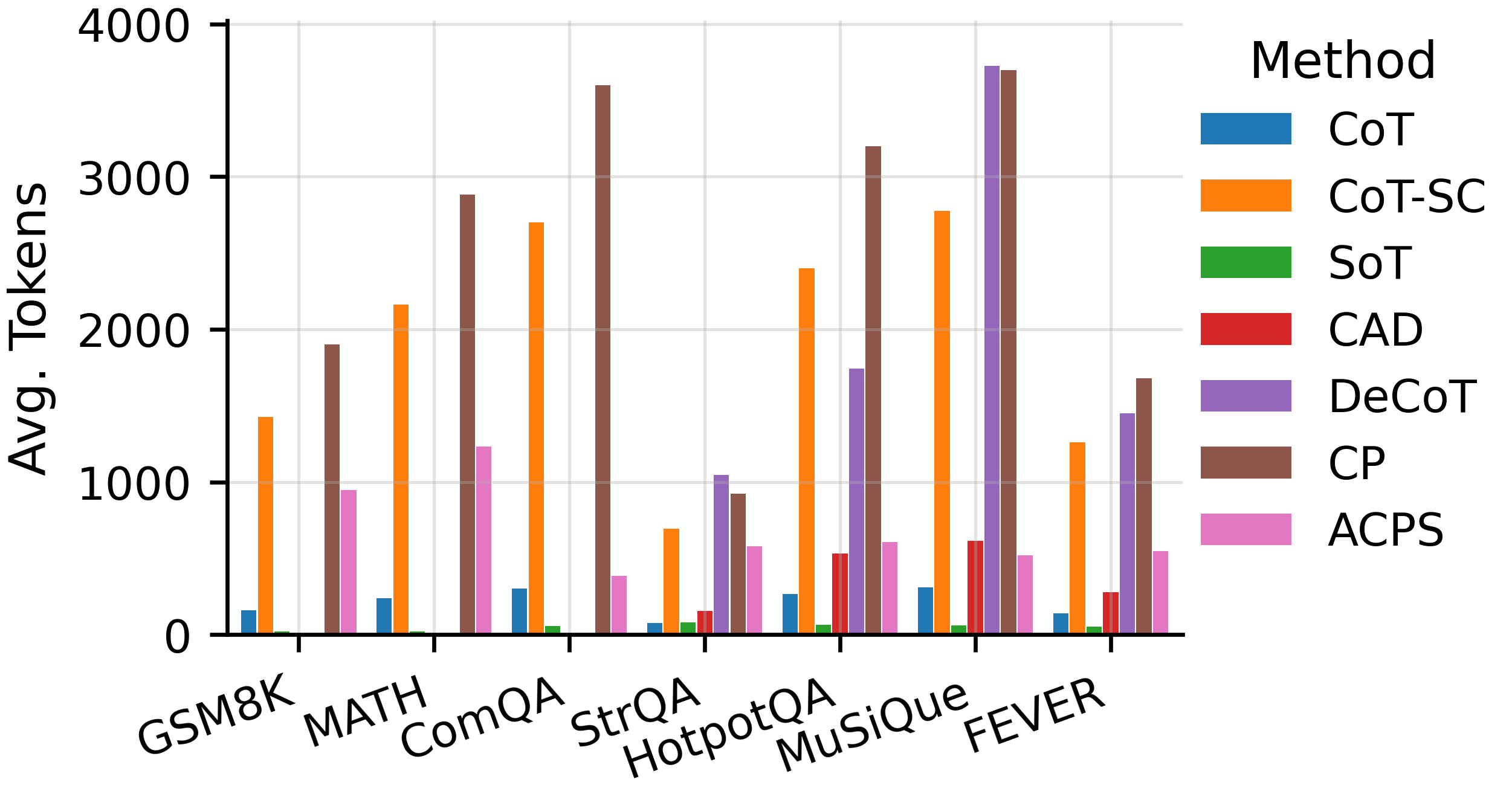}
    \caption{Comparison of average token consumed across all datasets for different prompting methods.}
    \label{fig:avg_token_ca}
\end{figure}

We analyse and compare the average tokens consumed by different prompting frameworks across multiple tasks, as shown in Figure~\ref{fig:avg_token_ca}. The results demonstrate that ACPS consistently requires fewer tokens while maintaining competitive performance, thereby demonstrating superior efficiency across diverse reasoning benchmarks.

\subsubsection{Accuracy-Efficiency Trade-off under Token Budgets}
\label{Accuracy-Efficiency}

\begin{figure}[t]
    \centering
    \includegraphics[width=0.40\textwidth]{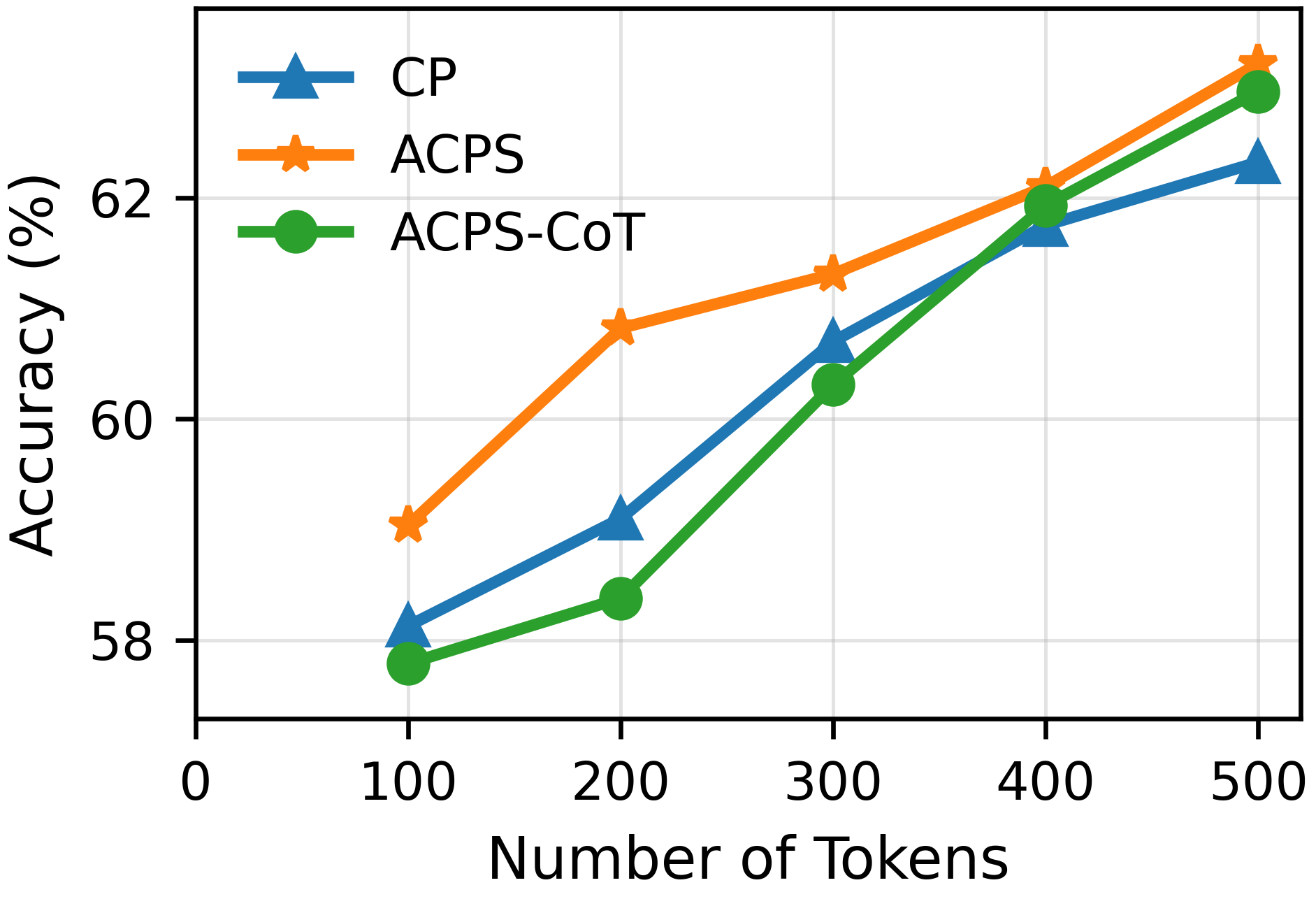}
    \caption{Comparison among CP, ACPS-CoT, and ACPS under varying token budgets on GPT-3.5-turbo for the HotpotQA dataset.}
    \label{fig:figure3}
\end{figure}

\begin{figure}[t]
    \centering
    \includegraphics[width=0.40\textwidth]{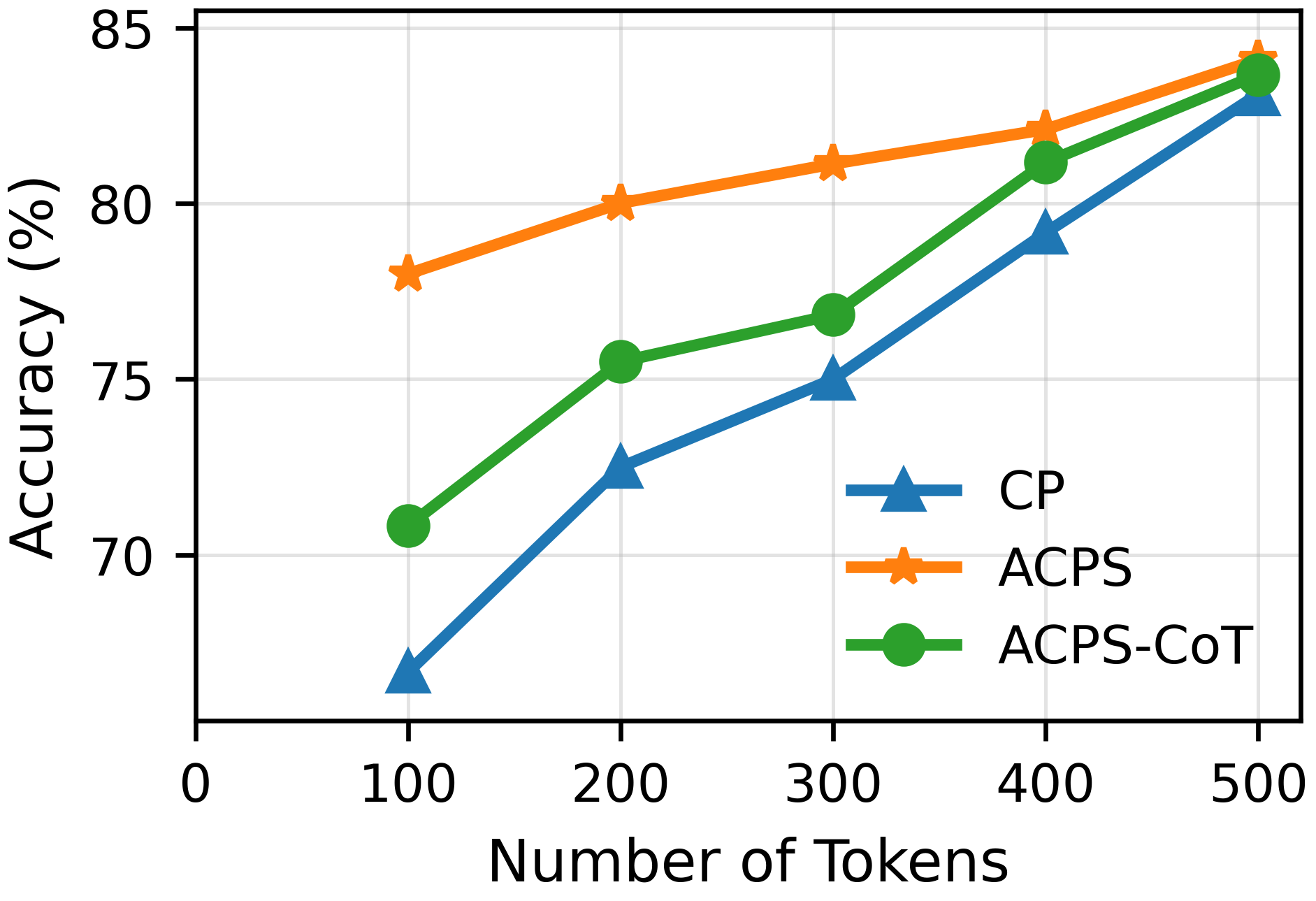}
    \caption{Comparison among CP, ACPS-CoT, and ACPS under varying token budgets on GPT-3.5-turbo for the StrategyQA dataset.}
    \label{fig:figure4}
\end{figure}

To examine the accuracy-efficiency trade-off under varying token budgets, we compare CP, ACPS, and ACPS-CoT on GPT-3.5-turbo using the StrategyQA and HotpotQA datasets, with \texttt{max\_tokens} varied up to 500. As shown in Figures~\ref{fig:figure3} and~\ref{fig:figure4}, ACPS consistently achieves the highest accuracy, demonstrating superior token efficiency. ACPS-CoT performs between CP and ACPS, yielding slightly better results than CP under the same token budgets but still falling short of ACPS. At comparable performance levels, ACPS requires fewer tokens than both CP and ACPS-CoT, further confirming its efficiency advantage.

\begin{table*}[t]
\centering
\small
\begin{tabular}{c|c|c|c|c|cc|cc|c}
    \toprule
    & {GSM8K} & {MATH}  & {ComQA} & {StrQA} 
    & \multicolumn{2}{c}{{HotpotQA}}
    & \multicolumn{2}{c|}{{MuSiQue}}
    & {FEVER} \\
\midrule
    {M}   & {Acc}~$\uparrow$ & {Acc}~$\uparrow$ & {Acc}~$\uparrow$ & {Acc}~$\uparrow$ & {EM}~$\uparrow$ & {F1}~$\uparrow$ & {EM}~$\uparrow$ & {F1}~$\uparrow$ & {Acc}~$\uparrow$ \\
    \midrule
    % Data rows here, currently empty
     4 & 78.58 & 47.14 & 74.09 & 80.04 & 59.12 & 73.37 & 75.67 & 65.11 & 75.15 \\
     6 & 79.00 & 47.89 & 74.55 & 81.01 & 60.41 & 75.44 & 75.84 & 65.55 & 76.64 \\
     8 & 79.25 & 48.21 & 74.75 & 83.06 & 61.11 & 75.59 & 76.21 & 66.92 & 78.23 \\
     10 & 81.52 & {48.36} & 74.95 & 84.12 & 62.23 & 75.49 & 77.03 & 67.21 & 79.84 \\
     12 & {82.30} & 47.65 & {75.10} & {84.32} & 61.56 & 75.69 & {77.89} & {67.94} & {80.11} \\\midrule\midrule
    {K}   & {Acc}~$\uparrow$ & {Acc}~$\uparrow$ & {Acc}~$\uparrow$ & {Acc}~$\uparrow$ & {EM}~$\uparrow$ & {F1}~$\uparrow$ & {EM}~$\uparrow$ & {F1}~$\uparrow$ & {Acc}~$\uparrow$ \\
    \midrule
    1 & 77.25 & 47.25 & 72.84 & 78.01 & 60.25 & 76.67 & 75.59 & 66.01 & 76.89 \\
    3 & 78.92 & 49.51 & 73.21 & 84.20 & 61.11 & 77.88 & 76.52 & 66.11 & 77.24 \\
    5 & {84.18} & 51.10 & 75.34 & 81.11 & 61.33 & 76.73 & 77.12 & 68.19 & 78.13 \\
    7 & 80.00 & 52.12 & 75.21 & 80.20 & 59.65 & 76.23 & 77.81 & 69.10 & 78.87 \\
    9 & 81.50 & {53.21} & {76.80} & 75.03 & 58.22 & 75.03 & {77.95} & {69.12} & {79.32} \\
    \bottomrule
\end{tabular}
\caption{The performance of ACPS under different numbers of generated SoTs (\(M\)) and clusters (\(K\)) across seven datasets.}
\label{tab:hyperparam_m}
\end{table*}

\subsection{Hyper-parameter Study}
\label{appendix:hyper-study}
We conduct a hyper-parameter study to examine how varying the number of initially generated SoTs (\(M\)) and the number of clusters (\(K\)) influences the performance of our framework. Due to computational constraints, we explore \(M \in \{4, 6, 8, 10, 12\}\) and \(N \in \{1, 3, 5, 7, 9\}\). The results are reported in Table~\ref{tab:hyperparam_m}. Overall, increasing \(M\) generally improves performance, as a larger and more diverse set of initial SoTs captures richer reasoning trajectories. Similarly, increasing \(N\) allows for finer-grained clustering of SoTs, which enhances the robustness of causal effect estimation. However, both higher \(M\) and \(N\) values lead to greater token consumption and computational overhead.

% ~\cite{Cheng2024Disentangled, Xu2024CFDICLR, Cheng2024CIV, Cheng2024LongitudinalIV, Cheng2023CIVDeep, Cheng2023CondIVRep}

\section{Related Work}
\label{related}
Large language models have shown impressive performance on various NLP tasks when provided with effective prompts. To avoid the high costs of scaling model size, researchers have developed prompt-based strategies that enhance reasoning without additional training. ICL enables models to learn from a few examples within the prompt~\cite{Brown2020GPT3}, while CoT prompting encourages step-by-step reasoning to improve multi-hop inference~\cite{Wei2022CoT}. To address answer variability, self-consistency decoding samples multiple reasoning paths and selects the majority answer~\cite{Wang2023SelfCons}. More recently, SoT generates concise reasoning sketches to improve efficiency across tasks~\cite{Aytes2025SketchOfThought}. 

Despite their effectiveness, these strategies primarily rely on correlational signals, i.e., selecting examples or reasoning paths based on majority voting. Such reliance can reinforce internal biases within LLMs, leading to unfaithful outputs. This highlights the need for causally grounded prompting strategies that can more reliably guide model reasoning.

\section{Prompt Templates}
\label{appendix:promtp_templates}
This section presents the prompt templates for SoT prompting, along with those used in conjunction with the NWGM approximation.

\subsection{SoT Prompting}

\makepromptbox{Chunked Symbolism Prompt Template}{
\textbf{Instruction}

You are a reasoning expert specializing in \textbf{Chunked Symbolism}, a cognitive reasoning technique that organizes numerical reasoning into structured steps. Your goal is to utilize chunked symbolism by representing information through \textbf{equations, variables, and step-by-step arithmetic}, using minimal words.

Chunked Symbolism is inspired by the cognitive science principle of \textbf{chunking}—the idea that humans process information more efficiently when grouped into meaningful units. Rather than solving problems in a free-form manner, Chunked Symbolism breaks down complex operations into smaller, structured steps.

\textbf{This method is particularly effective for:}
\begin{itemize}[leftmargin=*,noitemsep]
  \item Mathematical problems (arithmetic, algebra, physics, engineering)
  \item Symbolic reasoning (logic-based computations, formula derivations)
  \item Technical calculations (financial modeling, physics simulations, unit conversions)
\end{itemize}

\textbf{How to Apply Chunked Symbolism:}
\begin{enumerate}[leftmargin=*,noitemsep]
  \item Identify variables—extract relevant numerical values and define variables.
  \item Write equations—represent the solution using explicit mathematical formulas.
  \item Perform step-by-step computations—solve in small, logical steps, keeping each line clear.
  \item Label units—maintain consistent unit representation to prevent ambiguity.
  \item Final answer formatting—present the answer in the provided format for clarity.
\end{enumerate}

\textbf{Rules \& Directives:}
\begin{itemize}[leftmargin=*,noitemsep]
  \item Use equations \& variables; define variables before computation and always use explicit equations to represent reasoning.
  \item Avoid redundant text; do not restate the problem—go directly to calculations, and use minimal context only if it aids understanding.
  \item Apply step-by-step arithmetic; break operations into small, structured steps and ensure each line contains only one computation for clarity.
  \item \textbf{Output format:}
  \begin{quote}
  \texttt{<think>} \\
  \texttt{Let’s think through this step by step} \\
  \texttt{[stepwise equations, variables, and computations]} \\
  \texttt{</think>} \\
  \texttt{\textbackslash boxed\{[Final answer]\}}
  \end{quote}
  \item For multiple-choice, return the correct letter option in the box.\\
  Always use minimal words.
\end{itemize}

\textbf{Demonstration}
\begin{enumerate}[leftmargin=*,noitemsep]
  \item Q: The question is: [question]
  \item Let us think step by step.
  \item A: \texttt{<think>} \\
      \texttt{Let’s think through this step by step} \\
      \texttt{[equations, variables, computations]} \\
      \texttt{</think>} \\
      \texttt{\textbackslash boxed\{answer\}}
\end{enumerate}

\textbf{Test example:}
\begin{enumerate}[leftmargin=*,noitemsep]
  \item Q: The question is: [question]
  \item Let us think step by step.
  \item A: \texttt{<think>} \\
      \texttt{Let’s think through this step by step} \\
      \texttt{[equations, variables, computations]} \\
      \texttt{</think>} \\
      \texttt{\textbackslash boxed\{answer\}}
\end{enumerate}
}

\makepromptbox{Structured Concept Linking Prompt Template}{
\textbf{Instruction}

You are a reasoning expert specializing in \textbf{structured concept linking} by connecting essential ideas in a logical sequence. Your goal is to \textbf{extract key terms} and present reasoning in \textbf{clear, stepwise chains} with minimal explanation.

This method integrates \textbf{associative recall} (direct lookups) and \textbf{multi-hop reasoning} (sequential dependencies) into a unified framework.

\textbf{This method is most effective for:}
\begin{itemize}[leftmargin=*,noitemsep]
  \item Commonsense reasoning (linking familiar ideas)
  \item Multi-hop inference (tracing logical or causal dependencies)
  \item Fact-based recall (retrieving knowledge with minimal cognitive load)
\end{itemize}

\textbf{How to Apply:}
\begin{enumerate}[leftmargin=*,noitemsep]
  \item Extract key concepts—identify the most relevant words or entities.
  \item Use minimal words—make each reasoning step concise and direct.
  \item Link steps sequentially—ensure clear and meaningful progression between concepts.
  \item Avoid full sentences—respond using structured keyword connections.
  \item Follow the required format—present answers as stepwise chains.
\end{enumerate}

\textbf{Rules \& Directives:}
\begin{itemize}[leftmargin=*,noitemsep]
  \item Use structured concept linking; each step must be logically connected (arrows \texttt{($\rightarrow$)} for dependencies).
  \item Avoid unnecessary text; do not restate the question or use full sentences.
  \item Maintain logical flow; concepts must be meaningfully ordered and contribute to the reasoning process.
  \item \textbf{Output format:}
  \begin{quote}
  \texttt{<think>} \\
  \texttt{Let’s think through this step by step} \\
  \texttt{[shorthand reasoning]} \\
  \texttt{</think>} \\
  \texttt{\textbackslash boxed\{[Final answer]\}}
  \end{quote}
  \item For multiple-choice, return the correct letter option in the box.\\
  For fact-based recall, return True or False in the box.\\
  Always use minimal words.
\end{itemize}

\textbf{Demonstration}
\begin{enumerate}[leftmargin=*,noitemsep]
  \item Q: The context is: [paragraphs]. The question is: [question]
  \item Let us think step by step.
  \item A: \texttt{<think>} \\
      \texttt{Let’s think through this step by step} \\
      \texttt{[shorthand reasoning]} \\
      \texttt{</think>} \\
      \texttt{\textbackslash boxed\{answer\}}
\end{enumerate}

\textbf{Test example:}
\begin{enumerate}[leftmargin=*,noitemsep]
  \item Q: The context is: [paragraphs]. The question is: [question]
  \item Let us think step by step.
  \item A: \texttt{<think>} \\
      \texttt{Let’s think through this step by step} \\
      \texttt{[shorthand reasoning]} \\
      \texttt{</think>} \\
      \texttt{\textbackslash boxed\{answer\}}
\end{enumerate}
}

\subsection{SoT Prompting with NWGM Approximation}

For the prompt template, we design a unified structure applicable across all reasoning paradigms, reflecting our goal of building a general-purpose framework that supports diverse tasks without task-specific engineering. The task type is dynamically inferred by the pre-trained model. Unlike prior work~\cite{zhang2024causal}, which requires prompting the LLM to return task-specific answer formats, our template instead guides the model to perform symbolic reasoning and derive answers with minimal token usage through ICL.

\makepromptbox{Common Prompt Template}{%
\textbf{Instruction}

You are a helpful assistant to perform [task type]. Based on the context, answer the question step by step and provide the final answer at the end. I will provide reasoning processes, and please improve them to ensure the correct answer.

\textbf{Demonstration}
\begin{enumerate}[leftmargin=*,noitemsep]
  \item Q: The question is: [question]
  \item Let us think step by step,
  \item The provided reasoning process is: [wrong\_cot]  
  \item A: The improved reasoning process is: [correct\_cot]
  \item Therefore, the correct answer is: [answer]
\end{enumerate}

\textbf{Test example:}
\begin{enumerate}[leftmargin=*,noitemsep]
  \item Q: The question is: [question]
  \item Let us think step by step,
  \item The provided reasoning process is: [r\_k]
  \item A: The improved reasoning process is: [improved\_rs]
  \item Therefore, the correct answer is: [answer]
\end{enumerate}
}

\section{Case Study}
\label{Case}
This section presents two illustrative examples from CommonsenseQA and HotpotQA, highlighting the intermediate outputs at each stage of the framework. For each raw reasoning path, three improved reasoning paths are generated; however, for brevity, only the most informative ones are shown.

\makepromptbox{Case Study on CommonsenseQA}{
\textbf{Question}

Where do adults use glue sticks?

\textbf{Choices}
\{A: classroom, B: desk drawer, C: at school, D: office, E: kitchen drawer\}

\textbf{Cluster Reasoning Paths}

\begin{itemize}[leftmargin=*,noitemsep]
  \item \textbf{Cluster 1 ($|C_0| = 3$, $P(r_0|do(X)) = 3/9 = 0.333$), $P(A = D \mid do(r_0)) = 0.333$}
    \begin{quote}\small
    \textbf{Raw reasoning path:} \texttt{<think> Let’s think through this step by step \#adults → office → desk </think> \textbackslash boxed\{office\}} \\
    \vspace{1mm}
    
    \textbf{Improved reasoning path 1:} \texttt{<improved\_rs> \#glue\_stick → \#adults → \#office </improved\_rs> \textbackslash boxed\{D\}} \\
    \vspace{1mm}
    \end{quote}

  \item \textbf{Cluster 2 ($|C_1| = 3$, $P(r_1|do(X)) = 3/9 = 0.333$), $P(A = D \mid do(r_1)) = 0.667$}
    \begin{quote}\small
    \textbf{Raw reasoning path:} \texttt{<think> Let’s think through this step by step \#glue\_stick → children → school \#adults → office </think> \textbackslash boxed\{office\}} \\
    \vspace{1mm}

    \textbf{Improved reasoning path 1:} \texttt{<improved\_rs> \#glue\_stick → \#adults → \#office </improved\_rs> \textbackslash boxed\{D\}} \\
    \vspace{1mm}
    \end{quote}

  \item \textbf{Cluster 3 ($|C_2| = 2$, $P(r_2|do(X)) = 2/9 = 0.222$), $P(A = C \mid do(r_2)) = 0.667$}
    \begin{quote}\small
    \textbf{Raw reasoning path:} \texttt{<think> Let’s think through this step by step \#glue\_stick → common\_use → school </think> \textbackslash boxed\{at school\}} \\
    \vspace{1mm}

    \textbf{Improved reasoning path 1:} \texttt{<improved\_rs> \#glue\_stick → \#common\_use → \#school </improved\_rs> \textbackslash boxed\{C\}} \\
    \vspace{1mm}
    \end{quote}

  \item \textbf{Cluster 4 ($|C_3| = 1$, $P(r_3|do(X)) = 1/9 = 0.111$), $P(A = A \mid do(r_3)) = 1.000$}
    \begin{quote}\small
    \textbf{Raw reasoning path:} \texttt{<think> Let’s think through this step by step \#adulthood → specific\_tasks → work\_supplies → desk\_drawer → classroom\_setting → glue\_stick\_use </think> \textbackslash boxed\{classroom\}} \\
    \vspace{1mm}

    \textbf{Improved reasoning path 1:} \texttt{<improved\_rs> \#glue\_stick → \#adult\_tasks → \#teaching → \#classroom </improved\_rs> \textbackslash boxed\{A\}} \\
    \vspace{1mm}
    \end{quote}

\end{itemize}

\textbf{Weighted Voting and Final Answer}

Each cluster’s answer is assigned a probability weight. The final answer is determined by summing these weights and selecting the answer with the highest total.

\begin{itemize}[leftmargin=*,noitemsep]
  \item $P(A = D|do(X)) = 0.333 \times 0.333 + 0.333 \times 0.667 = 0.3333$
  \item $P(A = C|do(X)) = 0.222 \times 0.667 = 0.1481$
  \item $P(A = A|do(X)) = 0.111 \times 1.000 = 0.1111$
\end{itemize}

The answer with the largest total weight is selected as the final answer.

\textbf{Final Answer:} \texttt{\textbackslash boxed\{D\}}
}

\makepromptbox{Case Study on HotpotQA}{
\textbf{Question}

The driver know for doing backflips off his car lost to which driver in the 2009 NASCAR Nationwide Series?

\textbf{Context}
With 25 top-five finishes, Kyle Busch was the season champion.  He finished 210 points clear of Carl Edwards and 318 ahead of Brad Keselowski.  Edwards is well known for doing a backflip off of his car to celebrate a victory, which was a result of saving himself from a potential fall when he had his first win.

\textbf{Cluster Reasoning Paths}

\begin{itemize}[leftmargin=*,noitemsep]

  \item \textbf{Cluster 1 ($|C_0| = 3$, $P(r_0|do(X)) = 3/8 = 0.375$), $P(A = \text{Kyle Busch} \mid do(r_0)) = 3/3 = 1.000$}
    \begin{quote}\small
    \textbf{Raw reasoning path:} \texttt{<think> Let’s think through this step by step \#Driver known for backflips off car → Edwards \#Edwards → lost to Kyle Busch in 2009 NASCAR Nationwide Series </think> \textbackslash boxed\{Kyle Busch\}} \\
    \end{quote}

  \item \textbf{Cluster 2 ($|C_1| = 2$, $P(r_1|do(X)) = 2/8 = 0.250$), $P(A = \text{Carl Edwards} \mid do(r_1)) = 2/3 = 0.667$}
    \begin{quote}\small
    \textbf{Raw reasoning path:} \texttt{<think> Let’s think through this step by step \#Driver known for doing backflips → Carl Edwards \#2009 NASCAR Nationwide Series winner → Kyle Busch \#Closest competitor to the winner → Carl Edwards (210 points behind) \#Driver who lost to the winner → Carl Edwards </think> \textbackslash boxed\{Carl Edwards\}} \\
    \textbf{Improved reasoning path 2:} \texttt{[improved\_rs] Let's think through this step by step: \#Driver known for backflips → Carl Edwards (he is known for doing a backflip off of his car to celebrate a victory) \#2009 NASCAR Nationwide Series winner → Carl Edwards (Edwards, not Kyle Busch, was the known driver for backflips) </improved\_rs> \textbackslash boxed\{Carl Edwards\}} \\
    \vspace{1mm}
    \textbf{Improved reasoning path 3:} \texttt{Improved reasoning for the test example: \#Driver known for doing backflips → Carl Edwards \#Driver who lost to 2009 NASCAR Nationwide Series winner → Driver who lost to Kyle Busch Since the context mentions that Kyle Busch was the season champion with Carl Edwards and Brad Keselowski being the closest competitors, it can be inferred that Carl Edwards is the driver who lost to Kyle Busch. \textbackslash boxed\{Carl Edwards\}} \\
    \vspace{1mm}
    \end{quote}

  \item \textbf{Cluster 3 ($|C_2| = 2$, $P(r_2|do(X)) = 2/8 = 0.250$), $P(A = \text{Kyle Busch} \mid do(r_2)) = 2/3 = 0.667$}
    \begin{quote}\small
    \textbf{Raw reasoning path:} \texttt{<think> Let’s analyze the information: \#Context mentions Carl Edwards is famous for backflips \#Question asks who the backflip driver lost to in 2009 \#Carl Edwards did not win the championship in 2009 \#Therefore, Carl Edwards is the driver referenced as losing in the question </think> \textbackslash boxed\{Carl Edwards\}} \\
    \vspace{1mm}
    \textbf{Improved reasoning path 1:} \texttt{<improved\_rs> Let’s think through this step by step \#Driver known for backflips off car → Edwards \#Edwards lost to → Kyle Busch </improved\_rs> \textbackslash boxed\{Kyle Busch\}} \\
    \vspace{1mm}
    % \textbf{Improved reasoning path 2:} \texttt{<improved\_rs> \#Driver celebrated victories with backflips → Carl Edwards \#Question asks for the driver known for backflips who lost in 2009 \#Therefore, the answer is Carl Edwards </improved\_rs> \textbackslash boxed\{Carl Edwards\}} \\
    % \vspace{1mm}
    \textbf{Improved reasoning path 3:} \texttt{<improved\_rs> Let’s think through this step by step \#Driver known for backflips off car → Edwards \#Edwards' competition in 2009 NASCAR Nationwide Series → Kyle Busch </improved\_rs> \textbackslash boxed\{Kyle Busch\}} \\
    \vspace{1mm}
    \end{quote}

  \item \textbf{Cluster 4 ($|C_3| = 1$, $P(r_3|do(X)) = 1/8 = 0.125$), $P(A = \text{Brad Keselowski} \mid do(r_3)) = 2/3 = 0.667$}
    \begin{quote}\small
    \textbf{Raw reasoning path:} \texttt{<think> Let’s think through this step by step \#Driver doing backflips → Carl Edwards \#2009 NASCAR Nationwide Series → won by Kyle Busch \#Driver who lost → not mentioned, but not Carl Edwards (who did backflips) </think> \textbackslash boxed\{not mentioned\}} \\
    \vspace{1mm}
    \textbf{Improved reasoning path 2:} \texttt{<improved\_rs> \#Driver doing backflips → Carl Edwards (from context) \#2009 NASCAR Nationwide Series → won by Kyle Busch \#Kyle Busch finished 318 points ahead of Brad Keselowski \#Driver who lost to the champion → Brad Keselowski </improved\_rs> \textbackslash boxed\{Brad Keselowski\}} \\
    \vspace{1mm}
    \textbf{Improved reasoning path 3:} \texttt{[improved\_rs] The improved reasoning path is: \#Driver doing backflips → Carl Edwards \#2009 NASCAR Nationwide Series → won by Kyle Busch \#Driver ranked 3rd → Brad Keselowski (the driver one step behind Carl Edwards, who is one step behind Kyle Busch) The correct answer is therefore: \textbackslash boxed\{Brad Keselowski\}} \\
    \vspace{1mm}
    \end{quote}

\end{itemize}

\textbf{Weighted Voting and Final Answer}

Each cluster’s answer is assigned a probability weight. The final answer is determined by summing these weights and selecting the answer with the highest total.

\begin{itemize}[leftmargin=*,noitemsep]
  \item $P(A = \text{Kyle Busch}|do(X)) = 0.375 \times 1.000 + 0.250 \times 0.667 = 0.5417$
  \item $P(A = \text{Carl Edwards}|do(X)) = 0.250 \times 0.667 = 0.1667$
  \item $P(A = \text{Brad Keselowski}|do(X)) = 0.125 \times 0.667 = 0.0833$
\end{itemize}

The answer with the largest total weight is selected as the final answer.

\textbf{Final Answer:} \texttt{\textbackslash boxed\{Kyle Busch\}}
}

\end{document}